%% file: neurips_2026.tex
\documentclass{article}

 \usepackage[preprint]{neurips_2026}

\usepackage[utf8]{inputenc} 
\usepackage[T1]{fontenc}    
\usepackage{hyperref}       
\usepackage{url}            
\usepackage{booktabs}       
\usepackage{amsfonts}       
\usepackage{nicefrac}       
\usepackage{microtype}      
\usepackage{xcolor}         

\usepackage{microtype}
\usepackage{graphicx}
\usepackage{subcaption}
\usepackage{booktabs} 

\usepackage{enumitem}
\usepackage{float}
\usepackage{hyperref}

\usepackage{amsmath}
\usepackage{amssymb}
\usepackage{mathtools}
\usepackage{amsthm}

\allowdisplaybreaks[4]

\usepackage{wrapfig}

\usepackage{placeins}

\usepackage{booktabs}
\usepackage{makecell}
\usepackage{graphicx}
\usepackage{amssymb} 
\usepackage{pifont}  
\definecolor{my_green}{RGB}{51,102,0}
\definecolor{my_red}{RGB}{204, 0, 0}
\definecolor{my_half}{RGB}{226,115,0} 
\renewcommand{\checkmark}{\textcolor{my_green}{\ding{51}}} 
\newcommand{\crossmark}{\textcolor{my_red}{\ding{55}}} 
\newcommand{\halfcheck}{\textcolor{my_half}{\ding{51}\rotatebox[origin=c]{-9.2}{\kern-0.7em\ding{55}}}}
\definecolor{paired-light-blue}{RGB}{198, 219, 239}
\definecolor{paired-dark-blue}{RGB}{49, 130, 188}
\definecolor{paired-light-orange}{RGB}{251, 208, 162}
\definecolor{paired-dark-orange}{RGB}{230, 85, 12}
\definecolor{paired-light-green}{RGB}{199, 233, 193}
\definecolor{paired-dark-green}{RGB}{49, 163, 83}
\definecolor{paired-light-purple}{RGB}{218, 218, 235}
\definecolor{paired-dark-purple}{RGB}{117, 107, 176}
\definecolor{paired-light-gray}{RGB}{217, 217, 217}
\definecolor{paired-dark-gray}{RGB}{99, 99, 99}
\definecolor{paired-light-pink}{RGB}{222, 158, 214}
\definecolor{paired-dark-pink}{RGB}{123, 65, 115}
\definecolor{paired-light-red}{RGB}{231, 150, 156}
\definecolor{paired-dark-red}{RGB}{131, 60, 56}
\definecolor{paired-light-yellow}{RGB}{231, 204, 149}
\definecolor{paired-dark-yellow}{RGB}{141, 109, 49}  
\definecolor{myblue}{RGB}{218,232,252}
\definecolor{mygray}{RGB}{220,220,220}
\definecolor{mypink}{RGB}{251,49,153}

\usepackage[capitalize,noabbrev]{cleveref}

\theoremstyle{plain}

\theoremstyle{definition}

\theoremstyle{remark}

\title{CV-Arena: An Open Benchmark for Instructional Computer Vision Problem Solving with Human-AI Collaborative Preferences}

\author{%
Fangzhou Lin$^{1,2,3}$, Peiran Li$^1$, Lingyu Xu$^2$, Wenjing Chen$^1$, Qianwen Ge$^4$, Shuo Xing$^1$,\\
\bf Mingyang Wu$^1$, Xiangbo Gao$^1$, Siyuan Yang$^1$, Kazunori Yamada$^3$, Ziming Zhang$^2$,\\
\bf Haichong Zhang$^2$, Zhen Dong$^{5,6}$, Ming-Hsuan Yang$^7$, Zhengzhong Tu$^{1\star}$
\\[2pt]
$^1$Texas A\&M University\quad $^2$Worcester Polytechnic Institute\quad $^3$Tohoku University\\
$^4$Georgia Institute of Technology\quad
$^5$NVIDIA \quad 
$^6$UCSB
\quad$^7$UC Merced
\\[2pt]
\small $^\star$Corresponding Author: \texttt{tzz@tamu.edu}.\\[2pt]
\textbf{\textcolor{magenta}{Project Website}}: \href{https://ark1234.github.io/cv-arena}{\color{black}{\texttt{ https://ark1234.github.io/cv-arena}}}
}

\begin{document}

\maketitle

\input{sec/0_abstract}

\input{sec/1_intro}

\input{sec/1_5_related}
\input{sec/2_cv_arena}
\input{sec/3_elo_judge}
\input{sec/4_cv_agent}
\input{sec/5_exp}
\bibliographystyle{plain}
\bibliography{references}

\appendix

\input{sec/related_suppl}
\input{sec/cv_arena_suppl}

\input{sec/X_suppl}

\end{document}

%% file: sec/0_abstract.tex
\begin{abstract}

Instruction-guided image editing is becoming a general interface for visual work, yet existing benchmarks still focus largely on narrow appearance edits and do not fully capture the diversity of real-image tasks in professional workflows.
Here, we define instructional computer vision problem solving as a broader formulation of image editing: given a real input image and a natural-language instruction, a system must produce an edited output that realizes the requested transformation while satisfying explicit preservation, geometric, physical, and usability constraints.
We introduce CV-Arena, an open benchmark designed to evaluate this capability at professional scales.
CV-Arena contains 12K high-resolution real-image instruction pairs spanning 16 instruction-based visual task types, constructed using CogRetriever, a dual-track retrieval-and-curation pipeline that combines targeted web search, agentic query refinement, verification, and traceability.
To evaluate models at scale while preserving human fidelity, we propose Active Elo, a human-AI collaborative preference protocol that leverages CV-Judge, a logic-gated, multi-dimensional VLM evaluator, to reject clear failures and resolve high-confidence comparisons; and to route close, high-quality comparisons to expert raters.
Mixed human and AI supervision is then aggregated through reliability-weighted Elo updates.
Our comprehensive evaluation of 21 systems, including proprietary, open-source, and agentic models, on CV-Arena reveals persistent gaps in instruction adherence, physical reasoning, structural control, and fine-grained detail preservation.
We further develop CV-Agent, a lightweight agentic model that combines planning, editing, and verification, and demonstrate that closed-loop reasoning is a promising direction for professional-grade instruction-following visual editing.

\end{abstract}

%% file: sec/1_intro.tex
\section{Introduction}
\label{sec:intro}

A long-standing problem in modern computer vision (CV) is to \textit{modify an image according to human intent}.
Instruction-guided image editing offers a natural interface for this goal, that is, given an image and a natural-language instruction, a system is expected to change only what is requested while preserving the rest~\cite{magicbrush,ye2025imgedit}.
Recent multimodal generative models, including vision-language models (VLMs) and unified generative models~\cite{wang2024rl,zhang2024vision,guo2024large,wang2025vision,lu2024deepseek}, have made this interface increasingly practical.
Both proprietary systems~\cite{gpt-image-1,comanici2025gemini} and open models~\cite{lin2025uniworld,ye2025imgedit,wu2025qwenimagetechnicalreport} are now being used for everyday editing and increasingly complex visual workflows~\cite{manus_1_6,OpenAI2025ChatGPTagent,lin2025jarvisevo}.

However, most prior work~\cite{hqedit,ultraedit,varedit2025} still formulates instruction-guided editing as a relatively narrow set of appearance-centric or stylistic transformations, which only partially reflects the diversity of professional real-image workflows~\cite{ye2025imgedit,magicbrush,qian2025pico,bytemorph,ku2023imagenhub,peng2024dreambench++}.
We argue that this framing is too restrictive, and thereby define \textbf{instructional computer vision problem solving (iCVPS)} as a broader formulation of image editing.
For instance, an instruction may require a system to restore degraded content, enhance low-light or hazy images, recover faded text, manipulate object pose or geometry, insert objects with physically consistent lighting, or perform structure-preserving outpainting~\cite{luo2025visual,janjua2026grounding,wang2025adapting,zhou2025low,gu2025improving,zhang2026adaptive,cao2025instruction,song2026insert,jia2025compbench,zhang2025self}.
%
%
This broader view exposes a professional gap that is not well captured by existing editing benchmarks.
\underline{First}, many current systems can produce visually plausible images but often re-synthesize the input rather than faithfully modifying it, leading to unintended content changes and constraint violations~\cite{wang2025complexbench}.
\underline{Second}, existing evaluation protocols are unreliable for subtle, high-resolution comparisons, where small local artifacts, text errors, boundary inconsistencies, or geometric mistakes may determine whether an output is usable.
\underline{Third}, current benchmarks rarely stress-test the multi-domain workload required by professional workflows, as we defined above, including restoration, computational photography, physically grounded composition, semantic manipulation, typography recovery, and geometry-driven structural edits.~\cite{naveed2025comprehensive,achiam2023gpt,team2023gemini,TheC3}.

To close this gap, we introduce \textbf{CV-Arena}, an open benchmark that targets a diverse set of iCVPS tasks that naturally fit the image--instruction interface, spanning restoration and enhancement, computational photography, physically grounded composition, semantic manipulation, geometry and structural control, and typography recovery.
Crucially, CV-Arena focuses on high-resolution, in-the-wild images whose content and quality resemble those encountered in real visual workflows, making it more realistic than prior benchmarks~\cite{ye2025imgedit,magicbrush,bytemorph,ku2023imagenhub,peng2024dreambench++}, whose image sizes are mostly small (e.g., 512).
To construct the dataset at scale, we develop a text-initiated multimodal retrieval pipeline that converts professional editing intents into targeted web search, candidate discovery, verification, and traceable data records.
We further combine this agentic acquisition process with manual search and expert curation to address rare, difficult scenarios and reduce redundancy, resulting in the CV-Arena Dataset, which contains 12K open-domain iCVPS data across diverse high-resolution settings.


A second challenge is \emph{scalable and reliable evaluation} for CV-Arena.
Classical image quality metrics such as PSNR and SSIM~\cite{psnr,ssim} ignore instruction adherence and semantic preservation, while embedding-based metrics capture only partial signals for high-fidelity professional edits~\cite{clip}.
Recent benchmarks increasingly adopt {VLM-as-a-judge} for scalable scoring~\cite{ye2025imgedit,wu2025editreward}, but VLM judges can be brittle on subtle or near-tied comparisons, especially when correctness depends on fine local details.
Arena-style human preference evaluation~\cite{chiang2024chatbot} is more faithful, but costly, difficult to scale, and vulnerable to low-quality voting in crowdsourced settings~\cite{zhao2025challenges,jiang2024genai}.
To address these limitations, we propose \textbf{Active Elo}, a human-AI collaborative preference protocol that combines automated judging with selective expert supervision.
Our proposed \textbf{CV-Judge} first performs logic-gated, multi-dimensional evaluation to identify clear failures and high-confidence preferences; Active Elo then routes close, high-quality comparisons to expert raters and aggregates both human and AI decisions through reliability-weighted Elo updates.
This design concentrates human effort on the most informative cases while preserving scalable coverage across models, tasks, and high-resolution outputs, enabling stable comparison of both single-pass editors and agentic systems under fixed annotation budgets.
%

Beyond benchmarking existing editors, we also study whether agentic reasoning can improve iCVPS.
To this end, we build \textbf{CV-Agent}, a lightweight agentic baseline that decouples high-level instruction understanding, planning, and verification from low-level image manipulation.
The agent uses a strong editor as a tool, and wraps it with a closed-loop reasoning process that refines the edit and checks whether the output satisfies the instruction and constraints.
Although simple, this baseline helps validate an important finding of CV-Arena: many failures are not caused only by image generation quality, but by missing planning, constraint checking, and self-verification.
In summary, our contributions are:
\begin{itemize}[leftmargin=*, topsep=0pt, itemsep=0.5pt, parsep=0pt, partopsep=0pt]
\item \textbf{CV-Arena,} an open, professional-grade benchmark for iCVPS on real, high-resolution images, covering task families beyond appearance-centric editing while preserving native aspect ratios.

\item \textbf{Active Elo System,} a scalable human-AI collaborative preference protocol that combines a logic-gated multi-dimensional VLM evaluator with selective expert annotation and reliability-weighted Elo aggregation under constrained human budgets.

\item \textbf{CV-Agent,} a lightweight agentic baseline that decouples high-level planning and verification from low-level image manipulation, demonstrating that closed-loop reasoning can improve instruction following and constraint satisfaction in professional-grade visual editing.
\end{itemize}

\begin{table*}[]
    \centering
    \normalsize
    \caption{\textbf{Comparison of Existing iCVPS Benchmarks.} \#Size and \#Tasks represent the number of samples and editing types. Max Res. denotes the maximum resolution. 
    We also mark important attributes such as Real Image, Physics, Reasoning, Low Level, and Complex to compare the dataset diversity.
    The last column demonstrates the evaluation protocols used in the benchmark.
    }
    \vspace{-2mm}
    \label{tab:dataset_comparison}
    \resizebox{1\textwidth}{!}{
        \begin{tabular}{l|ccc|ccccc|c}
        \toprule
        \textbf{Dataset} & \textbf{\#Size} & \textbf{\#Tasks} & \textbf{\makecell[c]{Max Res.\\(px)$\uparrow$}} & \textbf{\makecell[c]{Real Image}} & 
        \textbf{\makecell[c]{Physics}} & 
        \textbf{\makecell[c]{Reasoning}} & 
        \textbf{\makecell[c]{Low Level}} & 
        \textbf{\makecell[c]{Complex}} &
        \textbf{\makecell[c]{Metrics}} \\
        \midrule
        
        MagicBrush~\cite{magicbrush}        & 10K   & 5  &  $500$          & \checkmark   & 
        \crossmark   &
        \crossmark   &
        \crossmark   &
        \checkmark   & L1, L2, CLIP, DINO\\
        
        InstructPix2Pix~\cite{instructp2p}  & 313K  & 4    & $512$          & \crossmark  & 
        \crossmark   &
        \crossmark   &
        \crossmark   &
        \crossmark   & CLIP\\
        
        HQ-Edit~\cite{hqedit}               & 197K  & 6    & ${\geq} 768$  & \crossmark  & 
        \crossmark   &
        \crossmark   &
        \checkmark   &
        \crossmark   & GPT  \\
        
        SEED-Data-Edit~\cite{seeddataedit}  & 3.7M  & 6    & $768$   & 
        \crossmark &  
        \crossmark   &
        \crossmark   &
        \crossmark   &
        \checkmark   & N/A \\

        UltraEdit~\cite{ultraedit}          & 4M    & 9    & $512$          & \crossmark & 
        \crossmark   &
        \crossmark   &
        \checkmark   &
        \crossmark   & L1, L2, CLIP, DINO  \\
        
        AnyEdit~\cite{yu2025anyedit}              & 2.5M  & 25   & $512$          &     \checkmark    & 
        \checkmark   &
        \checkmark   &
        \crossmark   &
        \checkmark   & L1, CLIP, DINO \\
        
        ImgEdit~\cite{ye2025imgedit}        & 1.2M  & 13   & $\geq 1280$   & \checkmark &  
        \crossmark   &
        \checkmark   &
        \crossmark   &
        \crossmark   & GPT \\
        \midrule
        \textbf{CV-Arena}                    & 12K  & 16 &  $\geq 2048$   & \checkmark &  
        \checkmark   &
        \checkmark   &
        \checkmark   &
        \checkmark   & GPT + Human \\
        \bottomrule
        \end{tabular}
    }
    \vspace{-5mm}
\end{table*}

%% file: sec/1_5_related.tex
\section{Related Work}
\label{sec:related_work}
\vspace{-0.4cm}
Real-world visual understanding has driven AI progress since MNIST and ImageNet~\cite{lecun1998mnist,deng2009imagenet,voulodimos2018deep,szeliski2022computer}, but while recognition-oriented tasks have largely saturated~\cite{elngar2021image,minaee2021image,zou2023object}, higher-tier objectives involving image realism, visual naturalness, and the plausibility of edits~\cite{theis2024makes,li2023towards,ye2025imgedit,chen2025opengpt} remain far from solved. Existing instructional editing benchmarks reflect this gap: object-insertion datasets such as iHarmony4~\cite{iharmony4} and ObjectDrop~\cite{orida,objectdrop} reduce the task to appearance harmonization or static placement, ignoring dynamic interactions with deformable media; semantic editing benchmarks such as MagicBrush~\cite{magicbrush} and RefCOCO-Edit~\cite{referringimageediting} conflate insertion, replacement, and reconstruction, blurring evaluation signals; and geometry-related edits in MagicBrush, InstructPix2Pix, and AnyEdit~\cite{magicbrush,instructp2p,yu2025anyedit} are entangled with appearance changes, while ImgEdit~\cite{ye2025imgedit} frames complexity through interaction length rather than structural constraints. Typography and UI recovery~\cite{qu2023exploring,fang2025recognition} are similarly under-served despite their importance in professional workflows. CV-Arena addresses these gaps with a geometry- and physics-aware task design that isolates dynamic interaction, semantic manipulation, structural transformation, and typography restoration as first-class categories on real, high-resolution images. A more thorough discussion of each line of work is provided in Appendix~\ref{app:extended_related_work}.

%% file: sec/2_cv_arena.tex
\section{CV-Arena Dataset}
\label{sec:dataset}
\begin{figure*}[!t]
    \centering
    \includegraphics[width=0.95\linewidth]{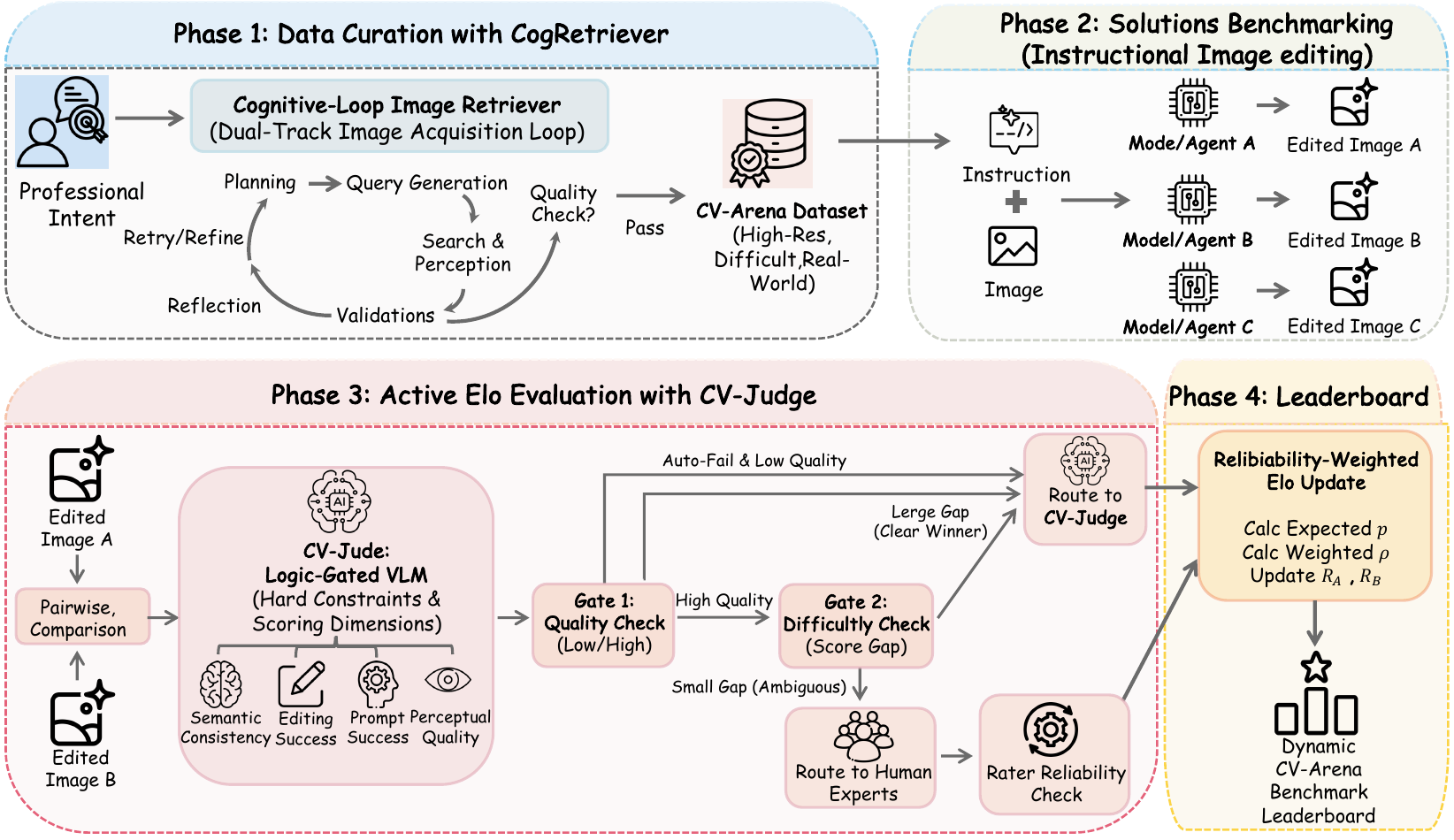}
    \vspace{-1mm}
    \caption{\textbf{The Overall Pipeline.} The framework starts with data curation, where CogRetriever constructs a professional-grade dataset. Then, it is followed by model/agent benchmarking and an Active Elo Evaluation, where CV-Judge generates scores and filters outputs using two-gate constraints, while routing ambiguous and high-quality comparisons to human experts. Final rankings are produced through Active Elo with a reliability-weighted update mechanism.}
    \label{fig:pipeline_arch}
    \vspace{-4mm}
\end{figure*}

Our dataset consists of 12k image-instruction pairs encompassing 16 distinct tasks. By integrating physical interaction and geometric constraints alongside traditional restoration tasks, CV-Arena spans the full spectrum from low-level pixel recovery to high-level structural manipulation. The construction follows a definition-driven pipeline (Figure~\ref{fig:pipeline_arch}): (i) defining instructional CV problem solving and deriving image selection criteria, (ii) designing a task taxonomy reflecting professional editing intents, and (iii) retrieving, filtering, and verifying real-world images for legality, quality, and traceability. Comparisons against concurrent datasets are summarized in Table~\ref{tab:dataset_comparison}.

\subsection{Problem Definition}
\label{sec:problem_def}

We formulate \textbf{instructional computer vision problem solving} (iCVPS) as a generalization of instruction-guided image editing. Given a real input image $x$ and a natural-language instruction $I$, a system must produce an edited output $\hat{x}=\mathrm{Edit}(x, I; m)$ that realizes the requested transformation while preserving everything that should remain unchanged.


This formulation introduces a set of professional constraints that go beyond perceptual realism: the output should additionally satisfy instruction adherence, semantic preservation, physical plausibility, geometric consistency, and high-resolution usability~\cite{luo2025visual,janjua2026grounding,wang2025adapting,zhou2025low,gu2025improving,zhang2026adaptive,cao2025instruction,song2026insert,jia2025compbench,zhang2025self}. These constraints in turn drive the image selection criterion: each pair must contain sufficient visual evidence for the task, a visible and unambiguous target region, and a clear success condition. We intentionally retain difficult real-world conditions (complex lighting, cluttered scenes, fine local structures, non-canonical viewpoints) so long as the source remains visually interpretable and the task intent unambiguous.

\subsection{Task Design and Taxonomy}
\label{sec:emerging_tasks_in_dataset}

\begin{figure*}[!h]
    \centering
    \includegraphics[width=0.95\textwidth]{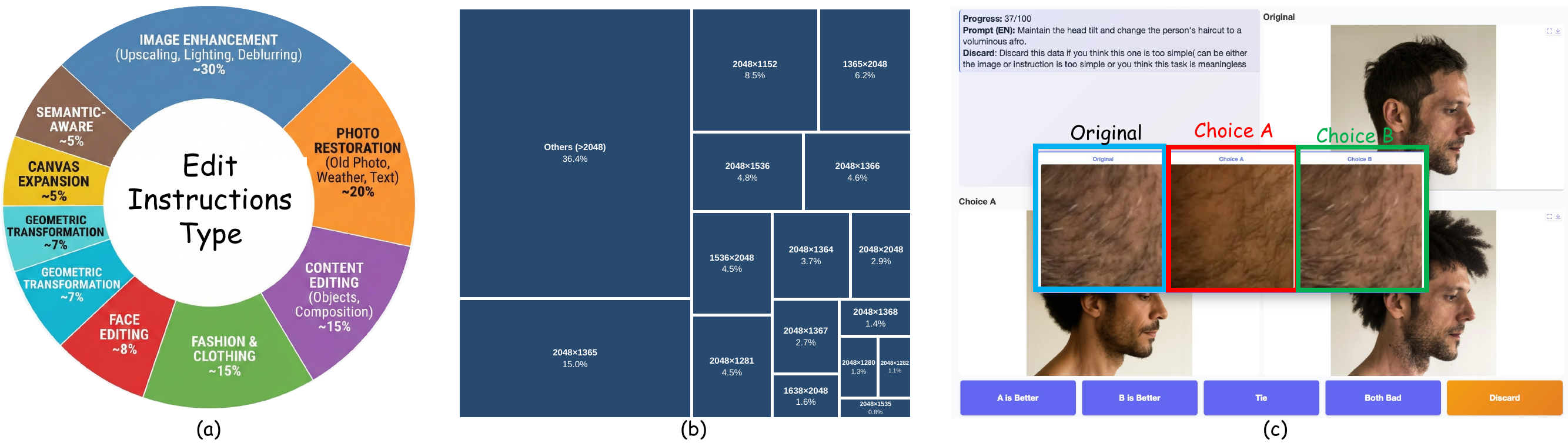}
    \vspace{-2mm}
    \caption{\textbf{Dataset statistics and User Interface.} From left to right: (a) the data composition across different sources; (b) shows the image-resolution
distributions; and (c) a \textit{zoom-in} function to check details during human rating.}
    \label{fig:statistics_key}
    \vspace{-1em}
\end{figure*}

The taxonomy is designed that collected images are guided by professional editing intents rather than organized post hoc. Beyond classical restoration tasks (exposure correction, deblurring, super-resolution) curated to reflect realistic difficulty, CV-Arena deliberately incorporates underrepresented task families critical to professional workflows: physically grounded scene composition, semantic-aware content manipulation, geometry-driven structural transformation, and typography or UI restoration in natural images. Figure~\ref{fig:statistics_key} (a, b) summarizes the task distribution and instruction keywords. Three task families particularly distinguish CV-Arena from prior benchmarks: \emph{Scene Composition and Object Insertion} requires physically and semantically coherent integration across geometry, lighting, scale, and semantics; \emph{Semantic-Aware Content Instruction} modifies intrinsic properties (pose, functional state, spatial configuration) without introducing or removing entities; and \emph{Text-Based Geometric Warping and Structural Control} performs precise, logically consistent shape transformations driven purely by language, including pose changes, viewpoint shifts, and fine-grained expression mixtures. Detailed task definitions are provided in Appendix~\ref{app:task_taxonomy}.

\subsection{Data Acquisition, Filtering, and Human Verification}

To collect images satisfying the above criteria at scale, we develop \textbf{CogRetriever}, a Text-Initiated Multimodal Search pipeline with a dual-track strategy. The \emph{Base Track} uses manual keyword search for high-precision acquisition in straightforward scenarios; the \emph{Agentic Track} provides scalable coverage for complex professional intents through a closed-loop system that maintains a reflection memory $m_t$ over $T$ iterations and operates in three stages: \textbf{\ding{182} Planning}, where a planner decomposes the instruction $\mathbf{I_i}$ into a diverse query set $\mathcal{Q}=\{q_1,\dots,q_K\}$ ($K=5$); \textbf{\ding{183} Action \& Perception}, where the system retrieves the top-$N$ candidates per query ($N=20$), validates them, and produces dense visual captions; and \textbf{\ding{184} Evaluation \& Pool Construction}, where a VLM scores candidates with $s(\mathbf{x};\mathbf{I_i})\in[0,1]$, retains those with $s\ge\tau$ ($\tau=0.8$), and writes a reflection $m_{t+1}$ if the pool fails to reach $K_p=3$ qualified samples. The final set is $\mathcal{X}^* = \mathrm{TopK}_{\mathbf{x} \in \mathcal{P}_t}\, s(\mathbf{x}; \mathbf{I_i})$. Full algorithmic details and hyperparameters are in Appendix~\ref{app:cogretriever}.

All retrieved images undergo automatic filtering for legality (Creative Commons rights filter for the Base Track; \texttt{cc\_publicdomain} and \texttt{cc\_attribute} restrictions via Google Custom Search API for the Agentic Track), near-duplicate removal, and low-quality rejection. Surviving pairs are then verified by human experts who check task-category match, target-region visibility, instruction feasibility, and the existence of a consistent success criterion, using the interactive \emph{zoom-in} tool (Figure~\ref{fig:statistics_key}c) for detail-sensitive cases. Complete filtering criteria, traceability logs, and the human-verification protocol are detailed in Appendix~\ref{app:filtering_verification}.

%% file: sec/3_elo_judge.tex
\section{Evaluation: Active Elo with CV-Judge}
\label{sec:evaluation}

We evaluate instructional computer vision problem-solving models through pairwise comparisons under identical conditions, aiming to obtain a reliable ranking rather than only absolute quality scores.
The bottom of Figure~\ref{fig:pipeline_arch} summarizes the overall evaluation pipeline.
Our evaluation stack consists of two components: \textbf{CV-Judge}, a multi-modal evaluation protocol for instructional image editing, and \textbf{Active Elo}, a human-AI collaborative ranking framework that allocates expert annotation to ambiguous comparisons and aggregates mixed supervision through reliability-aware updates.

\subsection{Preliminaries: Arena and Elo Ranking}
Arena-style evaluation has become a standard protocol for comparing open-ended generative systems~\cite{chiang2024chatbot}: rather than assigning absolute scores, annotators provide blinded pairwise preferences over two outputs produced under the same input, which is typically more stable when outputs are diverse and hard to calibrate on a universal scale. An Elo-style system then converts these wins and losses into a global leaderboard: each model $m$ maintains a rating $R_m$, and the win probability of $A$ over $B$ is a monotonic function following the Bradley-Terry-Luce model~\cite{bradley1952rank}. CV-Arena adopts this pairwise ranking view but modifies the standard protocol to account for expert annotation cost and the varying reliability of automatic judgments.

\subsection{Active Elo with Human-AI Collaboration}
\label{sec:hybrid_elo}


\textbf{Pairwise sampling.}
Let $\{(\mathbf{x_i}, \mathbf{I_i})\}_{i=1}^{N}$ denote the image-instruction pairs in CV-Arena.
For a model $m$, the edited output is
\begin{align}
\hat{x}_{i,m} = \mathrm{Edit}(\mathbf{x_i}, \mathbf{I_i}; m).
\end{align}
For any two models $(A,B)$ on the same instance $i$, we compare their outputs $\hat{x}_{i,A}$ and $\hat{x}_{i,B}$ under identical input conditions.
The evaluator produces a scalar score
\begin{equation}
s_{i,m} := \text{CV-Judge}\left(\mathbf{x_i} \mathbf{I_i}, \text{Edit}(\mathbf{x_i}, \mathbf{I_i}; m)\right),
\end{equation}
and induces a binary preference outcome
$z_{i,A,B}\in\{0,1\}$, where $z_{i,A,B}=1$ indicates $\hat{x}_{i,A}\succ \hat{x}_{i,B}$, i.e., $s_{i,A}\geq s_{i,B}$.
Human-routed pairs follow the same blinded pairwise format, but the final outcome is determined by expert preference rather than the automatic score difference.

\textbf{CV-Judge evaluation.}
Evaluating instructional image editing requires checking whether an edited image faithfully follows a given instruction while preserving the original image content that should remain unchanged.
Given an original image $a$, an instruction $I$, and an edited image $A$, CV-Judge operates original image, instruction and an edited image to produces a structured evaluation consisting of a scalar score, a binary success flag, and four auxiliary dimension scores retained for analysis and debugging.
The four dimensions are semantic consistency, editing success, prompt following, and perceptual quality.
\emph{Semantic consistency} measures whether identities, key objects, and layout not intended to change are preserved.
\emph{Editing success} captures whether the core edit specified by the instruction is actually realized with sufficient strength.
\emph{Prompt following} evaluates adherence to detailed instruction constraints, including explicit restrictions.
\emph{Perceptual quality} assesses visual realism and usability, penalizing artifacts, unnatural blending, or structural distortions.
Together, these dimensions disentangle the correctness of editing from perceptual appearance.

We denote the four dimension scores as $S_{\text{sem}}$, $S_{\text{edit}}$, $S_{\text{prompt}}$, and $S_{\text{perc}}$, respectively.
Each dimension is internally scored on $[0,1000]$, and the initial overall score is computed as a weighted sum:
\begin{align}
S_{\text{init}} =
\omega_s\,S_{\text{sem}} +
\omega_e\,S_{\text{edit}} +
\omega_i\,S_{\text{prompt}} +
\omega_p\,S_{\text{perc}}.
\end{align}
The weighting prioritizes correct realization of the instruction over purely perceptual improvements, preventing visually pleasing but incorrect edits from receiving high scores.
To enforce logical consistency, CV-Judge applies hard constraints: if the core edit is largely unsuccessful
($S_{\text{edit}} < \omega_e \cdot 1000$), the final score is capped at $(\omega_e+\omega_i)\cdot1000$ and marked unsuccessful; if semantic consistency or perceptual quality is severely degraded
($S_{\text{sem}} < \omega_s \cdot 1000$ or $S_{\text{perc}} < \omega_p \cdot 1000$), the score is capped at $(\omega_p+\omega_s)\cdot1000$ and marked unsuccessful.
An edit is considered successful only when all dimensions exceed moderate thresholds, ensuring both correctness and usability.
We denote the final score after the capping operation as $S$.
In our implementation, CV-Judge is instantiated with GPT-4o as the backbone VLM; cross-VLM sensitivity and dimension-weight sensitivity are reported in Appendix~\ref{app:vlm_sensitivity} and Appendix~\ref{app:hyper_sensitivity}, respectively.

\textbf{Human-AI routing.}
Pure VLM judging scales well but can be unreliable on subtle comparisons, while human judgments are high-fidelity but expensive.
We therefore use a two-gate routing policy to decide which pairs should be sent to human experts.
For a pair $(A,B)$ on instance $i$, define the score gap$g_{i}(A,B) = \left| s_{i,A} - s_{i,B} \right|$. We route the pair to human annotation iff
\begin{align}
\min(s_{i,A}, s_{i,B}) \ge \tau
\quad \text{and} \quad
g_{i}(A,B) < \Delta.
\end{align}
The \emph{quality gate} $\min(\cdot)\ge\tau$ avoids spending human budget on obvious failure regimes where both outputs are unusable.
The \emph{ambiguity gate} $g<\Delta$ targets cases where the automatic judge is least reliable and where additional supervision most improves the ranking.
Pairs that do not pass the routing condition are resolved automatically by CV-Judge.
Appendix~\ref{app:optimal_design} provides an information-per-cost interpretation of this routing rule.
Empirically, AI-human agreement rises monotonically with $g$, from $56.3\%$ at $g<50$ to $94.8\%$ at $g\ge 200$ (Appendix~\ref{app:agreement_strat}), confirming that the routing policy concentrates human effort where the VLM is least reliable.
The gate is also task-adaptive: deferral rates range from $46.8\%$ for geometry-driven warping down to $26.2\%$ for restoration (Appendix~\ref{app:per_dim_deferral}), showing that the policy automatically reallocates supervision to harder task families.

\textbf{Reliability-weighted Elo update.}
Each model $m$ maintains an Elo rating $R_m$, which is updated online after each pairwise comparison.
For a match between $A$ and $B$, we model the probability that $A$ beats $B$ following the Bradley-Terry-Luce model~\citep{bradley1952rank}:
\begin{align}
p_{AB} = \sigma\!\left(\frac{R_A - R_B}{S_{AB}}\right),
\end{align}
where $\sigma$ is the sigmoid function and $S_{AB}=\frac{s_{i,A}+s_{i,B}}{2}$ is an instance-dependent scale derived from the two CV-Judge scores.
To combine heterogeneous supervision, we downweight noisy outcomes with a credibility weight $\rho\in[0,1]$.
We model a rater with reliability $q\in[0,1]$ as producing the correct preference with probability $q$ and a random guess otherwise.
Given $(p_{AB}, z_{i,A,B})$, define
\begin{align}
\resizebox{0.55\linewidth}{!}{$
w =
\begin{cases}
p_{AB}, & z_{i,A,B}=1,\\
1-p_{AB}, & z_{i,A,B}=0,
\end{cases}
\quad
\rho = \frac{q\,w}{q\,w + (1-q)\tfrac12}.
$}
\end{align}
Human labels use $q\approx 1$, while AI-resolved matches use an instance-dependent reliability $q=q_{\mathrm{AI}}(g_i(A,B))$ calibrated on a small held-out set (Appendix~\ref{app:q_calib}).

We then update Elo ratings by
\begin{align}
& R_A \leftarrow R_A + K_r\,\rho\,\big(z_{i,A,B}-p_{AB}\big),\  R_B \leftarrow R_B - K_r\,\rho\,\big(z_{i,A,B}-p_{AB}\big),
\end{align}
where $K_r$ is rater-dependent.
We use a larger step size for human matches ($K_H$) and a smaller one for AI matches ($K_{AI}=\alpha K_H$).

This design leverages AI for scalability while preventing abundant but noisier AI supervision from overwhelming high-fidelity evidence.
Appendix~\ref{app:online_em} connects this update to a rater-aware BT mixture objective.

\textbf{Validation of the routing policy.}
We validate the two-gate routing policy through a budget-controlled ablation with a fixed number of expert comparisons ($B_H$).
Following an LMArena-style blinded pairwise protocol~\cite{chiang2024chatbot}, we construct a small, high-confidence human ground-truth test set $\mathcal{H}_{\text{test}}$ with 4 stable models, 8 curated task categories, and 10 expert annotators.
We evaluate each routing strategy by agreement with humans ($\mathrm{Acc}_H$)~\cite{ouyang2022training,zheng2023judging} and leaderboard stability, measured by bootstrap Spearman rank correlation $\rho_S$~\cite{kendall1938new,dubois2024length} and RankStd~\cite{dubois2023alpacafarm,jiang2024genai}. As shown in Table~\ref{tab:ablation_main}, the proposed two-gate routing policy substantially improves human consistency while producing a stable ranking.
Ablation details are provided in Appendix~\ref{app:ablations}.

\begin{wraptable}{r}{6cm}
\centering
\vspace{-0.9cm}
\tiny
\caption{\textbf{Validation of The Two-Gate Routing Policy.} Fixed human budget $B_H$. }
\vspace{-0.2cm}
\label{tab:ablation_main}
\begin{tabular}{l|ccc}
\toprule
Method & $\mathrm{Acc}_H \uparrow$ & $\rho_S \uparrow$ & RankStd $\downarrow$ \\
\midrule

Human-only   & 60.7\% & 0.68 & 38.5 \\
CV-Judge only  & 51.4\% & 0.63 & 23.2 \\
Quality-only gate  & 68.8\% & 0.81 & 27.9 \\
Ambiguity-only gate  & 73.2\% & 0.75 &  26.7\\
\textbf{Two-gate (Ours)}  & \textbf{82.6\%} & \textbf{0.94} & \textbf{22.3} \\

\bottomrule
\end{tabular}
\end{wraptable}

%% file: sec/4_cv_agent.tex
\section{CV-Agent: Simple Agentic Baseline}
\label{sec:agentic_baseline}

\vspace{-0.2cm}   

In addition to evaluating standalone editing models, we introduce a simple agentic editing baseline that decouples high-level reasoning from low-level image manipulation. The baseline is powered by strong LVLMs and off-the-shelf expert editors~\cite{google2025gemini25card,nano-banana-pro} and follows a lightweight ReAct-style loop~\cite{yao2022react}; it is modular and requires no additional supervision or task-specific tuning. Although deliberately minimal, CV-Agent serves as a paradigm-validating baseline; a per-stage module ablation isolating Understanding, Planning, and Closed-Loop Refinement is reported in Appendix~\ref{app:cvagent_ablation}. The pipeline proceeds in three stages:

\begin{figure*}[!t]
    \centering
    
        \includegraphics[width=1\linewidth]{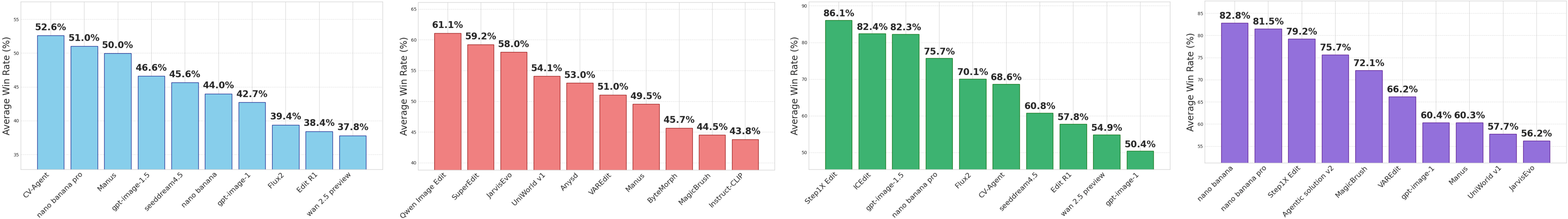}

    \caption{Average Win Rate Against Top-10 Models with three settings from left to right: Active Elo (Ours), Human Only, CV-Judge only, and EdiReward only(Assuming Uniform Sampling and No Ties).}
    \label{fig:avg_winrate}
    
    \vspace{1em} 

\includegraphics[width=1\linewidth]{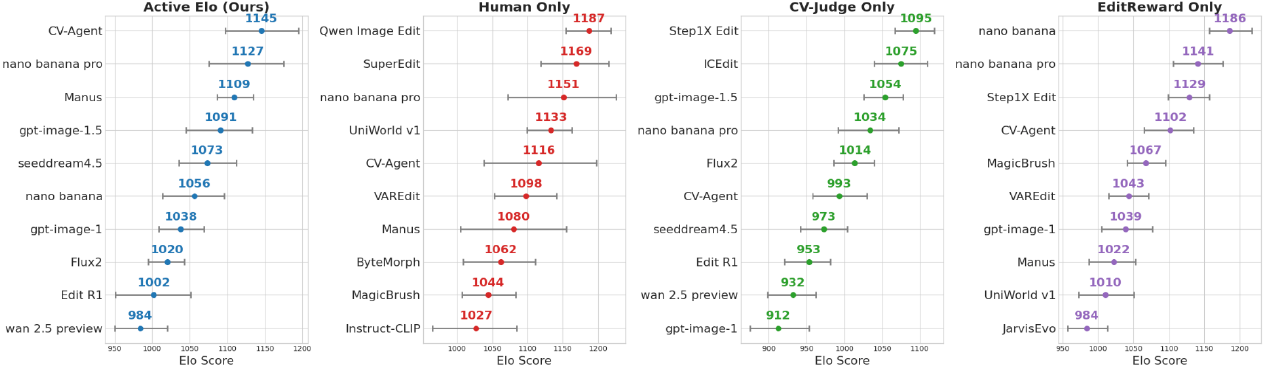}        

    \caption{Bootstrap of Elo Estimates (1000 Rounds of Random Sampling) on Top-10 Models with four settings from left to right: Active Elo (Ours), Human Only, CV-Judge only, and EdiReward only.}
    \label{fig:bootstrap}
\vspace{-0.5cm}    
\end{figure*}


\textbf{Stage\ding{182}: Understanding.}
Conditioned on $(\mathbf{x},\mathbf{I})$, the VLM (gemini-2.5-pro~\cite{google2025gemini25card}) rewrites $I$ into a precise, executable instruction and extracts the required visual changes and constraints. The output is a compact task specification that reduces ambiguity while preserving intent.

\textbf{Stage\ding{183}: Planning.}
The LVLM generates a structured plan. It also predicts whether the edit should be executed in one step or many, and sets a step budget capped by $T$ to prevent unbounded iteration.

\textbf{Stage\ding{184}: Closed-loop editing.}
For step $t$, the editor (nano banana pro~\cite{nano-banana-pro}) applies an edit to the current image using a step-specific prompt, producing $A_t$. The LVLM then evaluates $A_t$ against $(\mathbf{x},\mathbf{I})$ and outputs (i) a scalar quality score, (ii) a success indicator, and (iii) brief corrective feedback if needed. The loop stops early if the judge declares success; otherwise, it continues until $t=T$. The agent tracks the highest-scoring intermediate result and returns it as final output. 

\vspace{-0.2cm}

\begin{figure*}[]
    \centering
    \includegraphics[width=1\linewidth]{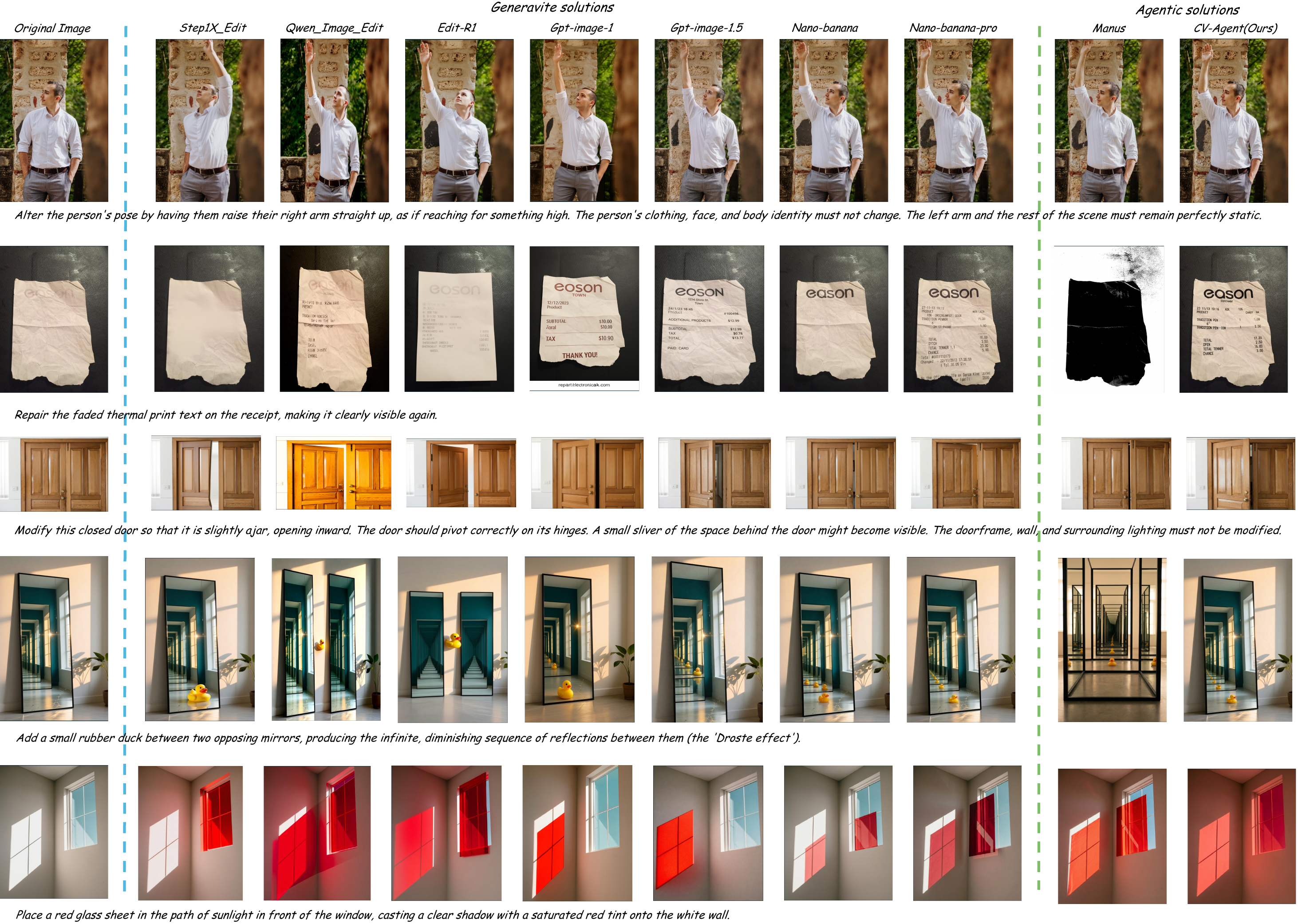}
    \vspace{-6mm}
    \caption{\textbf{Qualitative comparison among different editing solutions with reasoning and complex tasks}. Our proposed simple baseline CV-Agent consistently produces more faithful, constraint-satisfying edits, preserving non-target content and structure while better following the instruction than strong single-pass editors.}
    \label{fig:maincase}
\vspace{-0.65cm}
\end{figure*}

%% file: sec/5_exp.tex
\section{Experiments}
\label{sec:experiments}

\vspace{-0.2cm}

We benchmark a broad set of instructional image editing systems on CV-Arena, including both single-pass editors and agentic solutions. Our evaluation follows the \emph{Active Elo} introduced in Section~\ref{sec:hybrid_elo}, with three reference settings (CV-Judge only, Human-only~\cite{chiang2024chatbot,jiang2024genai}, and EditReward Only~\cite{wu2025editreward}) to isolate the effect of evaluation strategy. We first describe the evaluated models and experimental setup (Section~\ref{sec:eval_model}), and then report quantitative rankings (Table~\ref{tab:top10_comparison}) and qualitative analysis (Section~\ref{sec:eval_results}).


\subsection{Benchmark Details}
\label{sec:eval_model}


\textbf{Solutions}
We evaluate a diverse suite of instructional image editing solutions (\textit{21 solutions} in total). Closed-source systems include gpt-image-1~\cite{gpt-image-1}, gpt-image-1.5~\cite{gpt-image-1_5}, nano banana~\cite{nano-banana}, nano banana pro~\cite{nano-banana-pro}, Flux2~\cite{flux-2-2025}, and Seedream 4.5~\cite{seedream2025seedream},  wan 2.5 preview~\cite{wan2025}. Open-source baselines include Edit-R1~\cite{lin2025uniworld,ye2025imgedit},  VAREdit~\cite{varedit2025}, ICEdit~\cite{zhang2025icedit}, AnyEdit~\cite{yu2025anyedit}, Instruct-CLIP~\cite{chen2025instructclipimprovinginstructionguidedimage}, Step1X-Edit~\cite{liu2025step1x-edit}, MagicBrush~\cite{magicbrush}, UniWorld~\cite{lin2025uniworld}, Qwen Image Edit ~\cite{wu2025qwenimagetechnicalreport}, ByteMorph~\cite{chang2025bytemorph}, and SuperEdit~\cite{SuperEdit,MultiReward}. We also evaluate three agentic systems: Manus 1.6~\cite{manus_1_6}, JarvisEvo~\cite{lin2025jarvisevo}, and our CV-Agent.

\textbf{Inference protocol}
Editing is performed at each model's \emph{native output resolution} under their default/highest-quality inference setting. Unless otherwise stated, we do not apply post-processing that could alter the outputs, ensuring that measured differences reflect model behavior.

\textbf{Human evaluation interface}
Human comparisons are conducted on uniformly resized renderings for \emph{display only} to eliminate perceptual advantages from differing native resolutions; this does not affect any model output or any evaluation input to the judge. Importantly, because our protocol routes human primarily to \emph{high-quality and close} pairs, annotations often hinge on subtle local artifacts (e.g., detail legibility, boundary consistency, texture recovery). We therefore implement an interactive \emph{zoom-in} tool that allows humans to inspect fine details via mouse-controlled magnification (as shown in Figure~\ref{fig:statistics_key} c). This capability is critical for reliably differentiating near-tied outputs, and improves annotation fidelity in the regime our protocol explicitly targets.

\textbf{Evaluation settings}
We report three complementary evaluation settings:
(i) \textbf{Active Elo} (ours), as described in Section~\ref{sec:hybrid_elo}; (ii) \textbf{Human-only}, where comparisons are resolved by humans~\cite{chiang2024chatbot,jiang2024genai} (within budget); (iii) \textbf{CV-Judge only}, where all comparisons are resolved by our proposed automated judge, and (iv) \textbf{EditReward only}, where all comparisons are resolved by a concurrent automated judge~\cite{wu2025editreward}. These controlled references enable an apples-to-apples analysis of how hybrid routing and reliability-aware aggregation affect leaderboard quality and stability.


\subsection{Benchmark Results}
\label{sec:eval_results}

\vspace{-0.2cm}

\begin{table*}[t]
    \centering
    \small
    \caption{\textbf{Top-5 leaderboard comparison across evaluation settings.}
    We compare Active Elo (ours), human-only evaluation, CV-Judge-only evaluation, and EditReward-only evaluation.
    Active Elo combines scalable automatic judgments with selective expert supervision and aggregates mixed outcomes through reliability-weighted Elo updates.}
    \label{tab:top10_comparison}
    \vspace{-2mm}
    \resizebox{0.95\textwidth}{!}{%
    \begin{tabular}{lcc c lcc c lcc c lcc}
        \toprule
        \multicolumn{3}{c}{\textbf{Active Elo (Ours)}} & 
        & \multicolumn{3}{c}{\textbf{Human Only}} & 
        & \multicolumn{3}{c}{\textbf{CV-Judge Only}} & 
        & \multicolumn{3}{c}{\textbf{EditReward Only}} \\
        
        \cmidrule(lr){1-3}
        \cmidrule(lr){5-7}
        \cmidrule(lr){9-11}
        \cmidrule(lr){13-15}
        
        Model & Elo & 95\% CI &
        & Model & Elo & 95\% CI &
        & Model & Elo & 95\% CI &
        & Model & Elo & 95\% CI \\
        \midrule
        
        CV-Agent        & 1145 & +50/-48 & & Qwen Image Edit & 1187 & +31/-33 & & Step1X Edit      & 1095 & +24/-28 & & nano banana      & 1186 & +31/-29 \\
        nano banana pro & 1127 & +48/-52 & & SuperEdit        & 1169 & +46/-50 & & ICEdit           & 1075 & +35/-35 & & nano banana pro  & 1141 & +35/-35 \\
        Manus           & 1109 & +26/-22 & & nano banana pro  & 1151 & +75/-79 & & gpt-image-1.5    & 1054 & +24/-28 & & Step1X Edit      & 1129 & +28/-30 \\
        gpt-image-1.5   & 1091 & +42/-46 & & UniWorld v1      & 1133 & +30/-34 & & nano banana pro  & 1034 & +38/-42 & & CV-Agent         & 1102 & +33/-37 \\
        seeddream4.5    & 1073 & +39/-37 & & CV-Agent         & 1116 & +82/-78 & & Flux2            & 1014 & +26/-28 & & MagicBrush       & 1067 & +28/-26 \\
        
        \bottomrule
    \end{tabular}%
    }
    \vspace{-0.6cm}
\end{table*}

\textbf{Leaderboard.}
We rank solutions using the Human-AI Collaborative Preferences Active Elo protocol via credibility-weighted updates (Section~\ref{sec:hybrid_elo}). Our main leaderboard is shown in Table~\ref{tab:top10_comparison} (Please refer to the full 21 solutions in Table~\ref{tab:full_leaderboard} in the supplementary). In addition to our setting, we report the CV-Judge only, human-only, and EditReward Only leaderboard as a reference baseline. As shown in Table~\ref{tab:top10_comparison}, Human Only, CV-Judge, and EditReward alone cannot reflect the true competence of different solutions, whereas our Active Elo provides the most reliable ranking. We also include more results in supplementary, please refer Figure~\ref{fig:view_low_level} and Figure~\ref{fig:view_failure}.

\textbf{Discussion and Analysis.}
As shown in Figure~\ref{fig:avg_winrate}, we observe that the average win rates of the Top-10 ranked solutions in Active Elo and Human-only are similar, none exceeding 80\%. This indicates that there is no dominant, highly powerful solution in our CV-Arena Dataset at the current time. While CV-Judge and EditReward only clearly show untrustworthy results with exceeding 85\% in an open source model (Step1X Edit~\cite{liu2025step1x-edit}) or messed up ranking (nano banana~\cite{nano-banana} is better than nano banana pro~\cite{nano-banana-pro}), showing pure AI cannot handle our dataset. Moreover, we also included more analysis in Figure~\ref{fig:bootstrap}, the error bar of the Elo Score shows the reliability of Active Elo compared with Human Only, and is almost the same as CV-Judge only and EditReward only. As a result, our dynamic CV-Arena Benchmark more faithfully reflects human preference in the subtle, high-stakes regime that is most relevant for professional-grade instructional image editing. We also decompose CV-Judge scores along the four dimensions ($S_{\text{sem}}$, $S_{\text{edit}}$, $S_{\text{prompt}}$, $S_{\text{perc}}$) for the top solutions. Agentic methods (CV-Agent, Manus) lead on $S_{\text{edit}}$ and $S_{\text{prompt}}$, while strong single-pass generative models (nano banana pro, gpt-image-1.5, seeddream4.5) achieve higher $S_{\text{perc}}$. This separation suggests purely generative pipelines retain a perceptual edge that becomes decisive only when instruction adherence is otherwise comparable. Full per-dimension scores are reported in Appendix~\ref{app:per_dim_deferral}.

\textbf{Comparison with Traditional Metrics.} We also evaluate embedding-based similarity (CLIP-I~\cite{hafner2021clip}, DINO~\cite{zhang2022dino}) and text-image alignment (CLIPScore~\cite{hessel2021clipscore}) on a $\sim$1K subset. Top models cluster within a $3.4\%$ range under CLIP-I/DINO, and although paired bootstrap tests confirm many pairwise differences are statistically significant, the resulting ranking correlates only weakly with Active Elo (Spearman $\rho=0.50$) and produces rank reversals among competitive models, because input-output similarity rewards timid edits regardless of whether the instruction was actually realized. CLIPScore is substantially better aligned with human judgment ($\rho=0.90$) but still cannot resolve fine-grained perceptual artifacts and hard constraint violations that are decisive in our high-resolution professional setting. These observations support the use of traditional metrics cannot serve as a stand-alone leaderboard for this constraint-heavy task. Fullanalyses are in Appendix~\ref{app:traditional_metrics}.

\vspace{-0.3cm}



\section{Conclusion}
\label{sec:conclusion}
We introduce \textbf{CV-Arena}, an open benchmark designed to evaluate this capability at professional scales, together with \textbf{Active Elo}, a human-AI collaborative ranking protocol. Active Elo achieves substantially higher agreement with expert judgment and more stable leaderboards than VLM-only, reward-model-only, or budget-matched human-only baselines, and our simple \textbf{CV-Agent}, a neutral closed-loop agentic baseline to complete the benchmark and enable fair, end-to-end evaluation under a unified protocol.

\textbf{Limitations.}
The current 12K release, while sufficient for stable ranking, remains modest relative to the long tail of professional editing scenarios; scaling the dataset and broadening rare task coverage are left to future work.

%% file: sec/related_suppl.tex
\section{Extended Related Work}
\label{app:extended_related_work}
This appendix expands the related-work discussion summarized in Section~\ref{sec:related_work}, covering datasets and benchmarks for real-world visual understanding in greater detail.

\subsection{Datasets for Real-World Visual Understanding}

Since the early breakthroughs brought by the MNIST and ImageNet dataset~\cite{lecun1998mnist,deng2009imagenet}, real-world visual understanding has remained one of the most fundamental tasks in AI and has continuously driven the development of the entire AI community~\cite{voulodimos2018deep,forsyth2002computer,szeliski2022computer,zhang2024deep,zhao2024review}. With the rapid progress of modern architectures and large-scale training, many closed-form vision tasks have become largely saturated: such as image classification, semantic segmentation, and object detection~\cite{elngar2021image,minaee2021image,zou2023object}.
However, tasks at a higher semantic and perceptual tier, involving notions such as image realism~\cite{theis2024makes,chen2023hierarchical}, visual naturalness~\cite{li2023towards,chen2023exploring}, and the plausibility of edits~\cite{ye2025imgedit,chen2025opengpt}, are still far from being solved. Unlike recognition-oriented benchmarks, these tasks require modeling not only what is visible in the image, but also whether the visual content makes sense, remains natural, and aligns with implicit world knowledge and commonsense constraints.

\subsection{Benchmarks for Real-World Visual Understanding}

Existing benchmarks exhibit significant limitations in evaluating this capability. Early datasets, such as iHarmony4~\cite{iharmony4}, reduce object insertion to appearance harmonization, focusing primarily on color correction while ignoring essential physical cues such as shadows, reflections, and contact interactions. More recent counterfactual datasets~\cite{orida,objectdrop} improve realism by capturing static object presence on rigid surfaces, but still neglect dynamic interactions with deformable or non-solid media such as water, sand, or soft furniture. Benchmarks emphasizing placement plausibility~\cite{placement} further narrow the scope by evaluating only semantic appropriateness, overlooking physical consequences, aesthetic composition, and narrative coherence.

In CV-Arena, we shift the focus from static \emph{presence} to dynamic \emph{interaction}. We introduce scenarios probing interaction with deformable surfaces (e.g., ripples in water or footprints in sand), complex optical effects (e.g., distorted reflections or light caustics), and precise functional interactions (e.g., a key fitting into a lock).

Current benchmarks frequently conflate this task with simpler creation or erasure operations. Datasets such as MagicBrush~\cite{magicbrush} and RefCOCO-Edit~\cite{referringimageediting} mix reconstruction, replacement, and insertion tasks, blurring evaluation signals and obscuring true semantic understanding. CV-Arena isolates semantic manipulation as a first-priority task and emphasizes pose/state transition, spatial rearrangement, and component-level adjustments grounded in real images.

Existing benchmarks have partially touched upon geometry-related editing scenarios~\cite{magicbrush,instructp2p,yu2025anyedit}; however, such tasks are often not treated as a distinct category, or are represented by only a limited number of simplified cases. As a result, geometric transformation is frequently entangled with general appearance editing, making it difficult to isolate and evaluate a model’s structural reasoning capability. Some recent efforts, such as ImageEdit~\cite{ye2025imgedit}, explicitly introduce \emph{single-turn} and \emph{multi-turn} editing formulations to better support complex editing behaviors, partially addressing the limitations of one-shot editing protocols. While this design improves task coverage, it still frames complexity primarily from the perspective of interaction length rather than the underlying geometric constraints.

In contrast, CV-Arena adopts a geometry-centric task design that formulates editing scenarios based on the intended structural transformation itself, rather than explicitly categorizing tasks by the number of editing turns. It required editing process is implicitly determined by the geometric and structural complexity of the instruction, allowing tasks to more naturally reflect real-world professional workflows. This design choice not only evaluates a model’s image generation capability, but also probes its ability to accurately interpret and execute geometry-driven instructions, aligning more closely with the core objective of instruction-guided image editing.

Typography and UI restoration targets the correction, reconstruction, or removal of textual and graphical elements embedded in real-world images~\cite{qu2023exploring,fang2025recognition}. Real-world cases are substantially harder than synthetic overlays: restoring degraded signage, correcting distorted storefront typography, or removing watermarks/UI elements from faces or finely textured fabrics. These tasks require character accuracy, layout consistency, and seamless background integration.

CV-Arena explicitly incorporates typography- and UI-centric tasks (text in-painting/correction, watermark and complex graphic removal, layout-consistent restoration), reflecting professional standards beyond purely aesthetic outcomes.



%% file: sec/cv_arena_suppl.tex

\section{Appendix: Task Taxonomy Details}
\label{app:task_taxonomy}

This appendix provides the full definitions of the three signature task families that distinguish CV-Arena from prior benchmarks.

\paragraph{Scene Composition and Object Insertion.}
This category requires models to move beyond naive object pasting and perform physically and semantically coherent scene composition. Successful object insertion demands consistent integration across geometry, lighting, scale, and semantics, ensuring that inserted objects obey physical plausibility and visual harmony with the surrounding environment. Representative instructions include placing an object onto a surface with correct shadow casting, inserting a reflective object that respects the existing illumination, and composing multiple objects whose spatial arrangement must remain physically stable.

\paragraph{Semantic-Aware Content Instruction.}
Semantic-aware content instruction challenges a model to modify intrinsic properties of existing objects, such as pose, functional state, or spatial configuration, strictly without introducing or removing entities. Unlike object addition or deletion, these edits require fine-grained manipulation grounded in physical common sense and part-whole relationships. Representative instructions include changing the pose of an articulated object while preserving its identity, switching the functional state of a device (e.g., open versus closed), or rearranging spatial relationships among existing entities without altering the inventory of the scene.

\paragraph{Text-Based Geometric Warping and Structural Control.}
Text-based geometric warping requires models to perform precise, logically consistent shape and structure transformations driven purely by language. Representative instructions include pose transformations, viewpoint changes, and fine-grained expression control specifying continuous mixtures (e.g., ``slightly more surprised, less neutral'') rather than discrete categories. These tasks stress a model's ability to translate symbolic linguistic descriptions into geometrically faithful structural edits while preserving identity and surrounding context.

The remaining task families (restoration, enhancement, computational photography operations, typography and UI recovery, etc.) follow standard formulations from the literature but are curated at high resolution under realistic difficulty conditions.


\section{Appendix: CogRetriever Implementation Details}
\label{app:cogretriever}

\paragraph{Stage 1: Planning.}
Given a professional instruction $\mathbf{I_i}$, the planner decomposes $\mathbf{I_i}$ into searchable visual attributes and generates a diverse query set $\mathcal{Q}=\{q_1,\dots,q_K\}$, where $K=5$ is chosen to encourage coverage of complementary visual aspects (e.g., subject, scene, style, lighting, viewpoint).

\paragraph{Stage 2: Action \& Perception.}
For each query, the system searches and downloads the top-$N$ candidate images ($N=20$), applies multifaceted checks (file validity, minimum size, format normalization), and produces dense visual captions $c(x)$ that describe both semantic content and appearance attributes such as atmosphere, style, and composition. Captions are used both for downstream scoring and for filtering near-duplicates by content rather than by raw pixel similarity alone.

\paragraph{Stage 3: Evaluation \& Pool Construction.}
A vision-language model scores candidates with $s(\mathbf{x};\mathbf{I_i})\in[0,1]$, identifying a pool $\mathcal{P}_t = \{x \mid s(\mathbf{x};\mathbf{I_i})\ge\tau\}$. If $|\mathcal{P}_t| < K_p$ (where $K_p=3$, $\tau=0.8$), the agent writes a reflection in memory $m_{t+1}$ identifying missing or unexpected visual attributes, which guides query refinement in the next iteration. Once a sufficient pool is reached, the final candidate set is taken as the top-$K_p$ scoring elements:
\begin{equation}
\mathcal{X}^* = \mathrm{TopK}_{\mathbf{x} \in \mathcal{P}_t}\, s(\mathbf{x}; \mathbf{I_i}).
\end{equation}

\paragraph{Iteration cap and termination.}
The closed loop terminates either when $|\mathcal{P}_t|\ge K_p$ or when a maximum iteration budget $T$ is exhausted, in which case the instruction is flagged for manual review rather than silently producing low-quality samples.


\section{Appendix: Filtering, Traceability, and Human Verification}
\label{app:filtering_verification}

\paragraph{Comprehensive logging and traceability.}
To maintain dataset integrity and facilitate downstream auditability, the system records logs including the query set $\mathcal{Q}_t$, the evaluation scores $s_t$, and the corresponding agentic reflections at each iteration. The pipeline generates two outputs: (i) the finalized image pool selected for dataset inclusion, and (ii) a complete candidate set preserved with metadata for auditing or re-scoring. This logging design ensures reproducibility and supports rigorous offline analysis or re-evaluation under updated criteria.

\paragraph{Legality and copyright compliance.}
All imagery is retrieved through strictly filtered channels to ensure permissible usage. For manual acquisition in the Base Track, we use the \texttt{Creative Commons} rights filter in Google Images. The Agentic Track uses the Google Custom Search API with parameters restricted to \texttt{cc\_publicdomain} and \texttt{cc\_attribute} content.

\paragraph{Near-duplicate and low-quality removal.}
We additionally apply a filtering protocol that eliminates near-duplicates (using both perceptual hashing and caption similarity) and rejects low-quality samples (e.g., extreme compression artifacts, unrelated content). This stage further excludes ambiguous sources that might impede consistent evaluation. The filter is specifically designed to enhance the signal quality of the benchmark without oversimplifying the underlying tasks: we prioritize retention of challenging real-world conditions provided the source imagery remains visually interpretable and the associated task intent is unambiguous.

\paragraph{Human verification protocol.}
After automatic filtering, human experts further verify the remaining image-instruction pairs. The verification process checks four criteria: (i) whether the selected image matches the intended task category, (ii) whether the target region is visible, (iii) whether the instruction is feasible, and (iv) whether the expected edit can be judged consistently. For detail-sensitive cases, annotators use the \emph{zoom-in} function shown in Figure~\ref{fig:statistics_key}c to inspect local regions such as text, boundaries, object parts, and fine structural details. Pairs that fail any criterion are either re-routed to manual repair or discarded; the remaining examples constitute the final 12K dataset.

This final verification step ensures that the retained examples are legally traceable, visually interpretable, and aligned with the professional task taxonomy of CV-Arena.

%% file: sec/X_suppl.tex
\begin{table*}
    \centering
    \small 
    \caption{Full Leaderboard comparison across different settings (21 Models).}
    \label{tab:full_leaderboard}
    
    \resizebox{1\textwidth}{!}{%
    \begin{tabular}{lcc c lcc c lcc c lcc}
        \toprule
        \multicolumn{3}{c}{\textbf{Active Elo (Ours)}} & & \multicolumn{3}{c}{\textbf{Human Only}} & & \multicolumn{3}{c}{\textbf{CV-Judge only}} & & \multicolumn{3}{c}{\textbf{EditReward Only}} \\
        
        \cmidrule(lr){1-3} \cmidrule(lr){5-7} \cmidrule(lr){9-11} \cmidrule(lr){13-15}
        
        Model & Elo & 95\% CI & & Model & Elo & 95\% CI & & Model & Elo & 95\% CI & & Model & Elo & 95\% CI \\
        \midrule
        
        CV-Agent   & 1145 & +50/-48 & & Qwen Image Edit & 1187 & +31/-33 & & Step1X Edit & 1095 & +24/-28 & & nano banana & 1186 & +31/-29 \\
        nano banana pro & 1127 & +48/-52 & & SuperEdit & 1169 & +46/-50 & & ICEdit & 1075 & +35/-35 & & nano banana pro & 1141 & +35/-35 \\
        Manus & 1109 & +26/-22 & & nano banana pro & 1151 & +75/-79 & & gpt-image-1.5 & 1054 & +24/-28 & & Step1X Edit & 1129 & +28/-30 \\
        gpt-image-1.5 & 1091 & +42/-46 & & UniWorld v1 & 1133 & +30/-34 & & nano banana pro & 1034 & +38/-42 & & CV-Agent & 1102 & +33/-37 \\
        seeddream4.5 & 1073 & +39/-37 & & CV-Agent & 1116 & +82/-78 & & Flux2 & 1014 & +26/-28 & & MagicBrush & 1067 & +28/-26 \\
        nano banana & 1056 & +40/-42 & & VAREdit & 1098 & +43/-45 & & CV-Agent   & 993 & +37/-35 & & VAREdit & 1043 & +28/-28 \\
        gpt-image-1 & 1038 & +31/-29 & & Manus & 1080 & +75/-75 & & seeddream4.5 & 973 & +31/-31 & & gpt-image-1 & 1039 & +38/-34 \\
        Flux2 & 1020 & +23/-25 & & ByteMorph & 1062 & +49/-53 & & Edit R1 & 953 & +28/-32 & & Manus & 1022 & +31/-35 \\
        Edit R1 & 1002 & +49/-51 & & MagicBrush & 1044 & +39/-37 & & wan 2.5 preview & 932 & +30/-34 & & UniWorld v1 & 1010 & +41/-37 \\
        wan 2.5 preview & 984 & +36/-34 & & Instruct-CLIP & 1027 & +58/-62 & & gpt-image-1 & 912 & +41/-37 & & JarvisEvo & 984 & +29/-27 \\
        
        Qwen Image Edit & 965 & +36/-34 & & Step1X Edit & 1009 & +59/-61 & & nano banana & 892 & +35/-39 & & Anysd & 969 & +29/-25 \\
        VAREdit & 948 & +33/-37 & & JarvisEvo & 991 & +68/-70 & & Manus & 874 & +34/-32 & & gpt-image-1.5 & 952 & +36/-32 \\
        Step1X Edit & 932 & +30/-26 & & gpt-image-1.5 & 974 & +45/-41 & & SuperEdit & 856 & +27/-23 & & Edit R1 & 931 & +38/-34 \\
        UniWorld v1 & 915 & +38/-42 & & ICEdit & 956 & +42/-38 & & JarvisEvo & 839 & +42/-38 & & Instruct-CLIP & 909 & +29/-33 \\
        MagicBrush & 897 & +31/-27 & & Flux2 & 938 & +29/-31 & & Qwen Image Edit & 821 & +21/-25 & & SuperEdit & 894 & +25/-27 \\
        SuperEdit & 880 & +37/-37 & & seeddream4.5 & 920 & +70/-70 & & VAREdit & 804 & +18/-22 & & seeddream4.5 & 873 & +32/-36 \\
        ByteMorph & 864 & +30/-28 & & Edit R1 & 902 & +75/-73 & & Anysd & 787 & +37/-33 & & ByteMorph & 845 & +30/-34 \\
        ICEdit & 847 & +41/-39 & & gpt-image-1 & 884 & +78/-74 & & UniWorld v1 & 770 & +26/-28 & & Flux2 & 823 & +37/-37 \\
        Instruct-CLIP & 831 & +19/-23 & & nano banana & 866 & +74/-76 & & MagicBrush & 754 & +25/-27 & & ICEdit & 794 & +39/-39 \\
        Anysd & 815 & +35/-39 & & Anysd & 848 & +64/-64 & & ByteMorph & 738 & +32/-28 & & Qwen Image Edit & 771 & +30/-30 \\
        JarvisEvo & 799 & +32/-32 & & wan 2.5 preview & 830 & +43/-43 & & Instruct-CLIP & 722 & +36/-32 & & wan 2.5 preview & 756 & +35/-37 \\
        \bottomrule
    \end{tabular}%
    }
\vspace{-1mm}    
\end{table*}

\section{Appendix: Calibrating AI Reliability from Score Gap}
\label{app:q_calib}

Our hybrid protocol relies on an instance-dependent reliability for AI-resolved comparisons, denoted by $q_{\mathrm{AI}}(g)\in[0,1]$, where the score gap
\begin{align}
g_i(A,B) = |s_{i,A} - s_{i,B}|
\end{align}
serves as a practical proxy for comparison ambiguity. 

\subsection{Calibration set construction}

We construct a small calibration set of paired comparisons by sampling instances $i$ and model pairs $(A,B)$. For each sampled pair, we collect:
\begin{itemize}
  \item CV-Judge scores $(s_{i,A}, s_{i,B})$ and the induced AI preference
  \begin{align}
  \hat{z}^{\mathrm{AI}}_{i,A,B} = \mathbb{I}[s_{i,A} \ge s_{i,B}],
  \end{align}
  \item a human preference label $z^{\mathrm{H}}_{i,A,B}\in\{0,1\}$ under the same display protocol as the main benchmark.
\end{itemize}
We then define the agreement indicator
\begin{align}
a_{i,A,B}=\mathbb{I}\big[\hat{z}^{\mathrm{AI}}_{i,A,B}=z^{\mathrm{H}}_{i,A,B}\big]\in\{0,1\}.
\end{align}

\subsection{Binned estimation and monotone fitting}

We partition the observed gaps $\{g_{i}(A,B)\}$ into $J$ bins $\{ \mathcal{B}_j \}_{j=1}^J$ (e.g., equal-count bins for robustness). We compute the empirical AI reliability as the empirical probability of whether the AI preference agrees with human preference. In particular, the empirical AI reliability in bin $j$ is
\begin{align}
\hat{q}_j = \frac{1}{|\mathcal{B}_j|}\sum_{(i,A,B)\in\mathcal{B}_j} a_{i,A,B}.
\end{align}
Since reliability should be non-decreasing with $g$, we enforce monotonicity via either:
\begin{itemize}
  \item \textbf{Piecewise-constant monotone projection:} apply isotonic regression on $(\bar{g}_j,\hat{q}_j)$;
  \item \textbf{Smooth parametric form:} fit a logistic mapping

  \begin{equation} q_{\mathrm{AI}}(g) = \sigma\big(\beta (g - g_0)\big) \end{equation}
  
  where $\beta>0$ controls the sharpness of the transition, and $g_0$ denotes the ambiguity threshold, aligned with the routing criterion used in Section~\ref{sec:hybrid_elo}.
\end{itemize}
In our implementation, we default to isotonic regression for its nonparametric stability.

\subsection{Final reliability map used in ranking}

The calibrated function $q_{\mathrm{AI}}(g)$ is used in the credibility weight $\rho$ in the Elo updates (Section~\ref{sec:elo_update}). For AI-resolved comparisons we set
\begin{align}
q = q_{\mathrm{AI}}\big(g_i(A,B)\big),
\end{align}
while for human-labeled comparisons we use $q \approx 1$ (practically, $q=1-\varepsilon$ with a small $\varepsilon$ for numerical stability).

\section{Appendix: Two-Gate Selection as Cost-Effective Experimental Design}
\label{app:optimal_design}

We provide an interpretation of the two-gate routing rule as an approximate cost-effective design choice: allocate scarce human budget to comparisons that are both (i) relevant to the benchmark objective (high-quality regime) and (ii) most informative for refining the ranking (ambiguous regime).

\subsection{Information is concentrated in ambiguous comparisons}

Consider an online ranking step comparing models $A$ and $B$. Under Elo, the predicted win probability of $A$ is
\begin{align}
\label{eqn:p_AB_def}
p_{AB} = \sigma\!\left(\frac{R_A-R_B}{S}\right),
\end{align}
{where $\sigma$ is the sigmoid function, $S=\frac{s_{i,A}+s_{i,B}}{2}$}, which follows the BTL model. 
The informativeness of a single comparison can be measured by the variance of the Bernoulli outcome:
\begin{align}
\mathrm{Var}(A\text{ is ranked over B}\mid p_{AB})=p_{AB}(1-p_{AB}).
\end{align}
This variance is maximized when $p_{AB}=0.5$ (a toss-up) and decreases toward zero as $p_{AB}$ approaches 0 or 1. Intuitively, observing the outcome of a close match provides more information for refining the ranking than observing a heavily favored model win as expected.

Our routing rule leverages the CV-Judge score gap
\begin{align}
g_i(A,B)=|s_{i,A}-s_{i,B}|
\end{align}
as a proxy for comparison ambiguity. Smaller gaps indicate harder judgments for the automatic judge and greater uncertainty in the ordering. In such cases, human supervision provides the highest marginal benefit for improving ranking accuracy.

\subsection{Mixture model for noisy pairwise labels}

We adopt the observation model in Section~\ref{sec:experiments}. In particular,
 We introduce a latent variable $c\in\{0,1\}$ indicating whether the observed label is \emph{credible} ($c=1$) or a random guess ($c=0$). Given rater reliability $q\in[0,1]$, we assume
\begin{align}
P(c=1)=q,\qquad
P(z=1\mid c=1)=p_{AB},\qquad
P(z=1\mid c=0)=\tfrac12.
\end{align}
This yields the marginal observation model, i.e., 
\begin{align}
P(\mathbb{I}(A\succ B) \mid p_{AB}, q)= q\,p_{AB} + (1-q)\tfrac12.
\end{align}

matching Section~\ref{sec:experiments}. Human labels correspond to $q\approx 1$, while AI labels use $q=q_{\mathrm{AI}}(g)$.

\subsection{Rater reliability and effective information per cost}

Let us define $p_{\mathrm{eff}}=q\,p_{AB}+(1-q)/2$. The uncertainty of the observed label is governed by its variance $p_{\mathrm{eff}}(1-p_{\mathrm{eff}})$, while the \emph{informativeness} about $p_{AB}$ is attenuated when $q$ is small, since $p_{\mathrm{eff}}$ moves toward $1/2$ regardless of $p_{AB}$. Intuitively, if the judge is near-random on a subset of cases (low $q$), labels from that judge contribute little useful signal.

Let $c_{\mathrm{AI}}$ and $c_{\mathrm{H}}$ denote the per-comparison costs for AI and human supervision, respectively. A cost-effective design allocates human labels to cases where the expected gain in ranking quality per unit cost is larger. A simple proxy criterion is:
\begin{align}
\text{prefer human if}\quad
\frac{\text{info}(q_{\mathrm{H}},p_{AB})}{c_{\mathrm{H}}}
>
\frac{\text{info}(q_{\mathrm{AI}}(g(A,B)),p_{AB})}{c_{\mathrm{AI}}},
\end{align}
where $\text{info}(q, p_{AB})=(q\,p_{AB}+(1-q)/2)((1+q)/2-q\,p_{AB})$ increases with both ambiguity (near $p_{AB}=0.5$) and rater reliability.

Because $q_{\mathrm{H}}\approx 1$ and $q_{\mathrm{AI}}(g(A,B))$ decreases as $g$ becomes small (Appendix~\ref{app:q_calib}), human labeling becomes comparatively more valuable precisely in the ambiguous regime. This motivates the ambiguity gate $g(A,B)<\Delta$.

\subsection{Why the quality gate is necessary}

The benchmark objective emphasizes professional-grade competence, and low-quality outputs often lead to low-information human outcomes (e.g., ``both unusable'') that do not refine fine-grained ordering among competitive systems. We therefore restrict human effort to the regime where both candidates are at least moderately viable:
\begin{align}
\min(s_{i,A}, s_{i,B}) \ge \tau.
\end{align}
This can be viewed as multiplying the information-per-cost objective by a relevance indicator $u_i=\mathbb{I}[\min(\cdot)\ge\tau]$, effectively focusing annotation budget on comparisons aligned with the benchmark's evaluation regime.

\section{Appendix: Online-EM Interpretation of Reliability-Weighted Elo}
\label{app:online_em}

We provide an interpretation of the credibility-weighted Elo update as stochastic optimization of a rater-aware mixture objective. This is an \emph{interpretation} that explains why the weight $\rho$ is a principled way to combine heterogeneous supervision; the benchmark itself only requires the update rule in Section~\ref{sec:evaluation}.

\subsection{Posterior credibility}

Let us denote the observed outcome $z:=\mathbb{I}(A\succ B)$, and that  $z\in\{0,1\}$, define
\begin{align}
w=
\begin{cases}
p_{AB}, & z=1,\\
1-p_{AB}, & z=0.
\end{cases}
\end{align}
By Bayes' rule, the posterior probability that the label was generated from the credible component is
\begin{align}
\rho
:=P(c=1\mid z,p_{AB},q):&=\frac{P(c=1, z\mid p_{AB},q)}{P(c=1, z\mid p_{AB},q)+P(c=0, z\mid p_{AB},q)}\notag\\
&=
\frac{q\,w}{q\,w+(1-q)\tfrac12}.
\end{align}
This is exactly the credibility weight used in our Elo updates.

\subsection{Weighted log-likelihood and stochastic updates}

Consider the conditional log-likelihood of the credible component for a single comparison:
\begin{align}
\ell(R_A,R_B)
=
z\log p_{AB} + (1-z)\log(1-p_{AB}).
\end{align}
A credibility-weighted objective corresponds to maximizing $\rho\,\ell(R_A,R_B)$ online. By combining the definition of $p_{AB}$ in \eqref{eqn:p_AB_def}, we can compute the derivative of $l$ as follows
\begin{align}
\frac{\partial \ell}{\partial (R_A-R_B)} = \frac{1}{S}\,(z-p_{AB}),
\end{align}
then update Elo ratings by
\begin{align}
R_A \leftarrow R_A + \eta\,\rho\,(z-p_{AB}),
\qquad
R_B \leftarrow R_B - \eta\,\rho\,(z-p_{AB}),
\end{align}
which matches the reliability-weighted Elo update in Section~\ref{sec:evaluation} (with $\eta$ corresponding to $K_r$). Rater-dependent step sizes ($K_H$ vs.\ $K_{AI}$) can be viewed as an additional control that caps the influence of noisier supervision sources.

This perspective clarifies why $\rho$ is preferable to using the score gap alone as a weight: $\rho$ jointly captures (i) rater reliability $q$ and (ii) match difficulty through $p_{AB}$, and therefore directly controls how much each observed outcome should move the online ranking.

\section{Appendix: Two-Gate Routing Policy Ablations}
\label{app:ablations}

We test whether our Active Elo design choices are \emph{necessary} for producing a faithful and stable leaderboard. We ablate (i) \emph{routing} (which pairs receive human labels), (ii) \emph{reliability modeling} (whether AI trust must be instance-dependent), and (iii) \emph{aggregation} (whether noisy supervision should be downweighted). Notation follows the main text: CV-Judge scores $s_{i,m}$, gap $g_i(A,B)=|s_{i,A}-s_{i,B}|$, binary preference $z\in\{0,1\}$, Elo ratings $R_m$, and
\begin{align}
p_{AB}=\sigma\!\left(\frac{R_A-R_B}{S}\right),\qquad
\rho=\frac{q\,w}{q\,w+(1-q)\tfrac12},\ \ 
w=\begin{cases}p_{AB},& z=1\\ 1-p_{AB},& z=0.\end{cases}
\end{align}
We follow standard pairwise practice and use binary preferences (no ties).

\subsection{Ablation Settings}
\label{app:abl_what}

\textbf{Human GT for Evaluation.} We construct a small but high-confidence human ground truth (GT) as a set $\mathcal{H}_{\text{test}}$ to evaluate ranking faithfulness. Following the standard~\cite{jiang2024genai} protocol (blinded pairwise comparisons under identical display conditions), we (i) select $4$ empirically stable models, (ii) curate $8$ task categories that yield consistent discrimination, and (iii) recruit $10$ expert annotators. After repeated checks for consistency, we aggregate human preferences with a human-only pairwise ranker (Elo/BT) to obtain a GT \emph{ranking} and associated \emph{scores}. GT is used only for evaluation.

\textbf{Routing (human budget allocation).}
Under a fixed human budget $B_H$, we compare:
\begin{itemize}
  \item \textbf{CV-Judge only:} $z=\mathbb{I}[s_{i,A}\ge s_{i,B}]$ for all pairs;
  \item \textbf{Human-only (budget-matched):} rank using only $B_H$ human comparisons;
  \item \textbf{Quality-only:} human if $\min(s_{i,A},s_{i,B})\ge\tau$ (sample within-region to match $B_H$);
  \item \textbf{Ambiguity-only:} human if $g_i(A,B)<\Delta$;
  \item \textbf{Two-gate (ours):} human iff $\min(s_{i,A},s_{i,B})\ge\tau$ and $g_i(A,B)<\Delta$.
\end{itemize}

We report metrics that directly reflect (i) \emph{faithfulness to human preference} and (ii) \emph{leaderboard stability}. We avoid reporting raw Elo values, which are scale-dependent and less interpretable.

\textbf{(1) $\mathrm{Acc}_H$ (Human-consistency / Agreement with Humans).}
Given the final Elo ratings, we predict the preferred model in each held-out comparison $(A_k,B_k)$ as $\mathbb{I}[R_{A_k}>R_{B_k}]$ and compute agreement with human labels~\cite{ouyang2022training,zheng2023judging}:
\begin{align}
\mathrm{Acc}_{H}
=
\frac{1}{|\mathcal{H}_{\text{test}}|}
\sum_{k\in\mathcal{H}_{\text{test}}}
\mathbb{I}\!\left[
\big(R_{A_k}>R_{B_k}\big)\iff \big(z^{\mathrm{H}}_k=1\big)
\right].
\end{align}
Higher $\mathrm{Acc}_H$ indicates that the learned ranking better matches human pairwise preferences on the comparisons.

\textbf{(2) Spearman correlation (Rank correlation).}
To quantify ranking consistency under resampling, we perform bootstrap resampling of comparisons (with replacement), recompute Elo for each bootstrap replicate $b$, and obtain a ranking $\pi^{(b)}$. Let $\pi^{(\mathrm{full})}$ denote the ranking from the full comparison set under the same protocol. We report the average Spearman correlation between bootstrap and full-data ranks~\cite{kendall1938new,dubois2024length}:
\begin{align}
\rho_S
=
\frac{1}{B}\sum_{b=1}^{B}
\mathrm{Spearman}\!\left(\pi^{(b)},\,\pi^{(\mathrm{full})}\right),
\end{align}
where $B$ is the number of bootstrap replicates. Larger $\rho_S$ indicates that the relative ordering of models is stable under finite supervision.

\textbf{(3) Rank Std (Standard deviation of ranks / Bootstrap stability).}
Let $r^{(b)}_m$ be the rank of model $m$ in bootstrap replicate $b$. The rank standard deviation for model $m$ is
\begin{align}
\mathrm{StdRank}(m) = \mathrm{Std}\big(\{r^{(b)}_m\}_{b=1}^{B}\big),
\end{align}
and we summarize stability by the average rank standard deviation across models:
\begin{align}
\mathrm{RankStd}
=
\frac{1}{|\mathcal{M}|}\sum_{m\in\mathcal{M}} \mathrm{StdRank}(m),
\end{align}
where $\mathcal{M}$ is the model set. Lower $\mathrm{RankStd}$ indicates a more stable leaderboard (less sensitivity to the specific sampled comparisons)~\cite{dubois2023alpacafarm,jiang2024genai}.




As we can see in Table~\ref{tab:ablation_main}, 
two-gate routing improves sample efficiency under fixed $B_H$; it also improves both human agreement and stability relative to a constant trust level.

\begin{figure*}[b]
    \centering
    \includegraphics[width=1\linewidth]{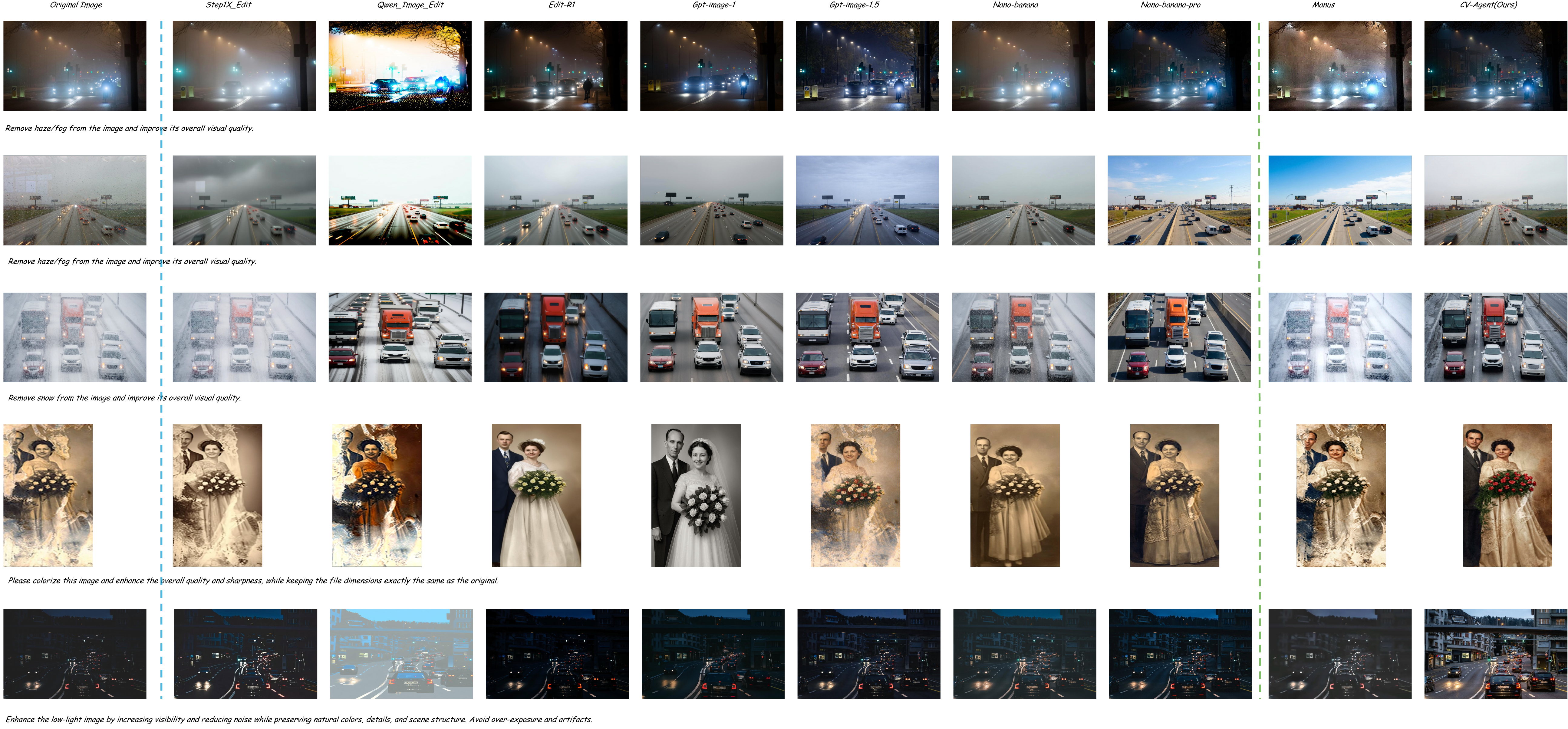}
    \caption{\textbf{Qualitative Comparison Among Different Editing Solutions with low level tasks}. }
    \label{fig:view_low_level}
\vspace{-3mm}
\end{figure*}

\begin{figure}[]
    \centering
    \includegraphics[width=0.5\linewidth]{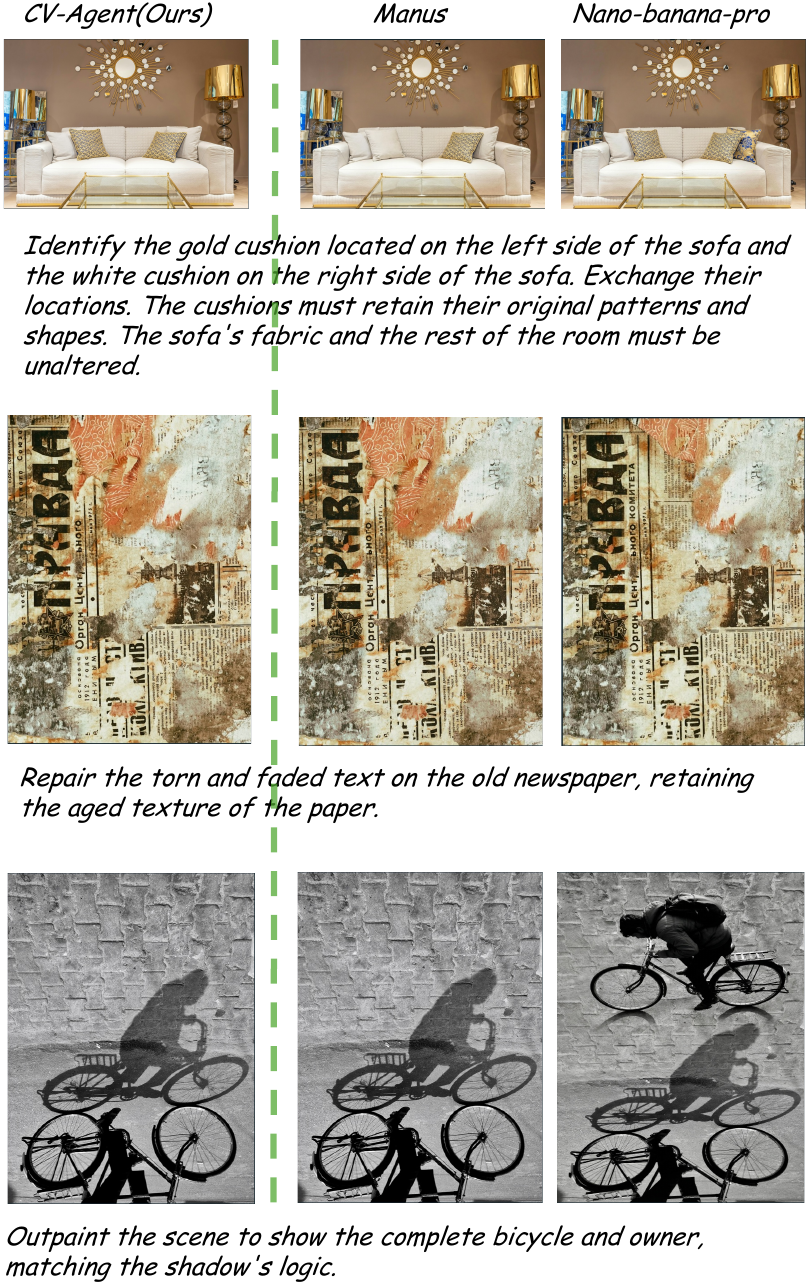}
    \caption{\textbf{Qualitative Comparison Among Different Editing Solutions with failure cases}.}
    \label{fig:view_failure}
\vspace{-3mm}
\end{figure}


\section{Appendix: CV-Judge VLM Backbone Sensitivity} 
\label{app:vlm_sensitivity}

CV-Judge is instantiated with GPT-4o as its backbone VLM, primarily due to the favorable balance between evaluation quality and API cost at the scale of CV-Arena. To verify that our findings are not artifacts of this specific choice, we conducted preliminary cross-VLM comparisons using alternative backbones from the GPT and Gemini families on a representative subset.

While per-category scoring distributions exhibit minor differences (notably for tasks requiring fine geometric reasoning), the overall Active Elo ranking remains largely stable across backbone choices. This robustness arises from three design choices working in concert: (i) the conservative quality gate operates only in the high-agreement regime ($g\ge 200$, $94.8\%$ AI-human agreement; Appendix~\ref{app:agreement_strat}); (ii) the ambiguity gate routes residual close-call cases to humans regardless of which backbone is used; and (iii) the reliability-weighted Elo update down-weights AI-resolved outcomes whose calibrated reliability is low. Together, these mechanisms localize backbone-specific biases to a small, gated portion of the budget rather than allowing them to dominate the leaderboard.


\section{Appendix: AI-Human Agreement Stratified by Score Gap}
\label{app:agreement_strat}

To diagnose where automatic CV-Judge decisions are reliable and where human supervision adds the most value, we stratify AI-human agreement by the score gap $g_i(A,B)=|s_{i,A}-s_{i,B}|$ on the held-out human GT set $\mathcal{H}_{\text{test}}$.

\begin{table}[h]
\centering
\small
\caption{AI-Human agreement and pair fraction by score gap.}
\label{tab:gap_strat}
\begin{tabular}{l|cc}
\toprule
Score gap $g$ & AI-Human agreement & Fraction of pairs \\
\midrule
$g<50$           & 56.3\% & 18.2\% \\
$50\le g<100$    & 69.1\% & 24.6\% \\
$100\le g<200$   & 83.5\% & 31.4\% \\
$g\ge 200$       & 94.8\% & 25.8\% \\
\bottomrule
\end{tabular}
\end{table}

Agreement increases monotonically with the gap. This validates two key design choices simultaneously: (i) the ambiguity gate $g<\Delta$ correctly identifies the regime where CV-Judge alone is unreliable and routes those cases to humans, and (ii) the AI reliability function $q_{\mathrm{AI}}(g)$ used in the credibility weight (Appendix~\ref{app:q_calib}) is well-calibrated, since AI-resolved outcomes outside the routing window already exceed $90\%$ agreement and can safely contribute to the Elo updates.


\section{Appendix: Per-Dimension Score Breakdown and Task-Level Deferral Rates} 
\label{app:per_dim_deferral}

\subsection{Per-Dimension Breakdown for Top Solutions}

We report mean CV-Judge scores (on $[0, 1000]$) along each evaluation dimension for the top-5 solutions under Active Elo.

\begin{table}[h]
\centering
\small
\caption{Per-dimension mean scores for top-5 Active Elo solutions.}
\label{tab:per_dim_table}
\begin{tabular}{l|cccc}
\toprule
Model & $S_{\text{sem}}$ & $S_{\text{edit}}$ & $S_{\text{prompt}}$ & $S_{\text{perc}}$ \\
\midrule
CV-Agent          & 782 & \textbf{798} & \textbf{785} & 751 \\
nano banana pro   & 756 & 741 & 732 & \textbf{784} \\
Manus             & 771 & 758 & 749 & 738 \\
gpt-image-1.5     & 748 & 726 & 731 & 769 \\
seeddream4.5      & 734 & 711 & 703 & 761 \\
\bottomrule
\end{tabular}
\end{table}

Two complementary patterns emerge: agentic solutions (CV-Agent, Manus) lead on $S_{\text{edit}}$ and $S_{\text{prompt}}$, while single-pass generative models (nano banana pro, gpt-image-1.5, seeddream4.5) achieve higher $S_{\text{perc}}$ but exhibit weaker instruction fidelity. This separation indicates that planning, verification, and closed-loop refinement most directly benefit instruction adherence, whereas purely generative pipelines retain a perceptual edge that is decisive only when instruction adherence is otherwise comparable.

\subsection{Task-Level Deferral Rates}

Approximately $33.7\%$ of pairwise comparisons satisfy both gates ($\min(s_{i,A},s_{i,B})\ge\tau$ and $g_i<\Delta$) and are routed to human annotators. The deferral rate, however, varies substantially across task families.

\begin{table}[h]
\centering
\small
\caption{Task-level human deferral rates.}
\label{tab:deferral_table}
\begin{tabular}{l|c}
\toprule
Task family & Deferral rate \\
\midrule
Geometry-driven warping            & 46.8\% \\
Physically grounded composition    & 44.6\% \\
Semantic content manipulation      & 33.5\% \\
Computational photography          & 28.7\% \\
Restoration / Enhancement          & 26.2\% \\
\bottomrule
\end{tabular}
\end{table}

Tasks that hinge on subtle constraint satisfaction (geometry, physics) generate more close-call comparisons and therefore consume proportionally more human budget; conversely, restoration tasks produce larger and more unambiguous quality gaps that CV-Judge resolves reliably. This adaptive allocation is an emergent property of the two-gate policy: human effort flows automatically toward task families where it is most informative, without any task-specific tuning.


\section{Appendix: Comparison with Traditional Metrics} 
\label{app:traditional_metrics}

For completeness, we evaluate three widely used reference-light metrics on a randomly sampled subset of $\sim$1K pairs: CLIP-I (input-output image similarity), DINO (visual feature similarity), and CLIPScore (text-image similarity). We caution at the outset that CV-Arena samples lack ground-truth edited references, so CLIP-I and DINO can only measure preservation of input content, which is not uniformly desirable across our 16 task families (e.g., object insertion legitimately deviates from the input, while exposure correction should preserve it).

\subsection{Aggregate Scores}

\begin{table}[h]
\centering
\small
\caption{Traditional metric scores on the $\sim$1K subset, with Active Elo rank for reference.}
\label{tab:trad_aggregate}
\begin{tabular}{l|cccc}
\toprule
Model & CLIP-I $\uparrow$ & DINO $\uparrow$ & CLIPScore $\uparrow$ & Active Elo \\
\midrule
CV-Agent         & 0.891 & 0.842 & 0.287 & 1 \\
Manus            & 0.876 & 0.831 & 0.281 & 3 \\
nano banana pro  & 0.864 & 0.817 & 0.272 & 2 \\
nano banana      & 0.869 & 0.823 & 0.256 & 6 \\
gpt-image-1.5    & 0.857 & 0.809 & 0.265 & 4 \\
\bottomrule
\end{tabular}
\end{table}

CLIP-I spans only $0.857$--$0.891$ (a $3.4\%$ range) and DINO spans $0.809$--$0.842$ ($3.3\%$). The narrow margins limit their power to discriminate competitive solutions.

\subsection{Paired Significance Testing}

We ran paired bootstrap tests ($B=10000$). With $N\approx 1000$, statistical power is high: CLIP-I yields $7/10$ significant pairs and DINO yields $6/10$. However, the resulting rankings are nearly identical across CLIP-I and DINO (CV-Agent $>$ Manus $>$ nano banana $>$ nano banana pro $>$ gpt-image-1.5) and Spearman-correlate only weakly with Active Elo ($\rho=0.50$, $p=0.39$). Two of the significant pairs are direct rank reversals relative to Active Elo, indicating that the disagreement is not statistical noise but systematic: input-output similarity rewards timid edits regardless of whether the instruction was actually realized.

\subsection{CLIPScore: Better but Still Insufficient}

CLIPScore correlates substantially better with Active Elo ($\rho=0.90$): top-1 and bottom-1 match exactly, with only a single adjacent swap among the middle ranks. This confirms that text-image alignment is a more faithful proxy for instruction adherence than input-output similarity. Nevertheless, CLIPScore's text encoder lacks the resolution to detect (i) fine-grained perceptual artifacts (boundary inconsistencies, texture corruption) and (ii) hard constraint violations (e.g., introducing an object the instruction explicitly forbids), both of which are decisive in professional-grade settings. Active Elo's multi-dimensional CV-Judge combined with selective human routing captures both axes that no single embedding-based metric covers, while still using these metrics as useful supplementary diagnostics.


\section{Appendix: Hyperparameter Sensitivity} 
\label{app:hyper_sensitivity}

\subsection{Routing Thresholds $\tau$ and $\Delta$}

The structural ablation in Table~\ref{tab:ablation_main} already isolates the role of each gate. We supplement it with a finer-grained sweep around the default operating point $(\tau^\star,\Delta^\star)$.

\begin{table}[h]
\centering
\small
\caption{Sensitivity of Active Elo to the routing thresholds.}
\label{tab:tau_delta}
\begin{tabular}{l|ccc}
\toprule
Configuration & $\mathrm{Acc}_H \uparrow$ & Spearman $\rho_S \uparrow$ & RankStd $\downarrow$ \\
\midrule
Default $(\tau^\star,\Delta^\star)$         & \textbf{82.6\%} & \textbf{0.94} & \textbf{22.3} \\
Loose $(\tau^\star-50,\Delta^\star+50)$     & 80.1\% & 0.91 & 23.6 \\
Tight $(\tau^\star+50,\Delta^\star-50)$     & 81.4\% & 0.92 & 22.8 \\
\bottomrule
\end{tabular}
\end{table}

Performance degrades gracefully under modest perturbations, confirming the protocol does not rely on knife-edge tuning.

\subsection{Dimension Weights $(\omega_s,\omega_e,\omega_i,\omega_p)$}

The dimension weights structurally determine both the weighted score and the hard constraint caps in CV-Judge. Our default $(0.20,0.30,0.30,0.20)$ prioritizes instruction correctness over purely perceptual appearance.

\begin{table}[h]
\centering
\small
\caption{Sensitivity to CV-Judge dimension weights.}
\label{tab:weights}
\begin{tabular}{l|cc}
\toprule
Configuration & $\mathrm{Acc}_H \uparrow$ & Spearman $\rho_S \uparrow$ \\
\midrule
Uniform $(0.25,0.25,0.25,0.25)$               & 78.1\% & 0.87 \\
\textbf{Default (ours)} $(0.20,0.30,0.30,0.20)$ & \textbf{82.6\%} & \textbf{0.94} \\
Edit-heavy $(0.15,0.40,0.30,0.15)$            & 79.8\% & 0.91 \\
\bottomrule
\end{tabular}
\end{table}

Uniform weighting over-rewards visually pleasing but instruction-violating outputs. Edit-heavy weights raise the edit-failure cap aggressively (to $\omega_e\times 1000=400$), which is too lenient: outputs that miss the edit by a small margin escape being flagged as failures. The default weighting occupies the operating point that best balances correctness against perceptual quality, and the top-5 ranking is stable across all three configurations.


\section{Appendix: CV-Agent Module Ablation} 
\label{app:cvagent_ablation}

CV-Agent is an intentionally minimal agentic baseline (Section~\ref{sec:agentic_baseline}) whose purpose is to validate the agentic paradigm rather than to demonstrate architectural novelty. We isolate the contribution of each stage in the three-stage pipeline.

\begin{table}[h]
\centering
\small
\caption{CV-Agent module ablation under the same Active Elo protocol.}
\label{tab:cvagent_ablation}
\begin{tabular}{l|c}
\toprule
Configuration & Active Elo \\
\midrule
Editor only (single pass)                              & 1056 \\
\hspace{1em}+ Stage 1 (Understanding)                  & 1089 \\
\hspace{1em}+ Stage 1 + Stage 2 (Planning)             & 1118 \\
\textbf{Full CV-Agent (Stages 1--3, closed-loop)}      & \textbf{1145} \\
\bottomrule
\end{tabular}
\end{table}

Each stage contributes a monotonic, non-trivial gain: instruction rewriting (Stage 1) improves the precision of low-level edits; planning (Stage 2) enables multi-step decomposition for complex requests; and closed-loop refinement (Stage 3) recovers the largest portion of remaining failures by detecting and correcting under-edits. Even with these very simple modules, the combined pipeline already achieves the top Active Elo position, supporting our claim that the agentic paradigm itself, rather than any specific architectural choice, is the primary driver of the observed improvement, and pointing to planning, verification, and closed-loop refinement as a promising direction for future work.

%% file: references.bib
@String(CVPR= {IEEE Conf. Comput. Vis. Pattern Recog.})

@String(AAAI = {AAAI})

@String(CVPR  = {CVPR})

@article{naveed2025comprehensive,
  title={A comprehensive overview of large language models},
  author={Naveed, Humza and Khan, Asad Ullah and Qiu, Shi and Saqib, Muhammad and Anwar, Saeed and Usman, Muhammad and Akhtar, Naveed and Barnes, Nick and Mian, Ajmal},
  journal={ACM Transactions on Intelligent Systems and Technology},
  volume={16},
  number={5},
  pages={1--72},
  year={2025},
  publisher={ACM New York, NY}
}

@article{achiam2023gpt,
  title={Gpt-4 technical report},
  author={Achiam, Josh and Adler, Steven and Agarwal, Sandhini and Ahmad, Lama and Akkaya, Ilge and Aleman, Florencia Leoni and Almeida, Diogo and Altenschmidt, Janko and Altman, Sam and Anadkat, Shyamal and others},
  journal={arXiv preprint arXiv:2303.08774},
  year={2023}
}

@article{team2023gemini,
  title={Gemini: a family of highly capable multimodal models},
  author={Team, Gemini and Anil, Rohan and Borgeaud, Sebastian and Alayrac, Jean-Baptiste and Yu, Jiahui and Soricut, Radu and Schalkwyk, Johan and Dai, Andrew M and Hauth, Anja and Millican, Katie and others},
  journal={arXiv preprint arXiv:2312.11805},
  year={2023}
}

@inproceedings{TheC3,
  title={The Claude 3 Model Family: Opus, Sonnet, Haiku},
  author={Anthropic PBC},
  url={https://api.semanticscholar.org/CorpusID:268232499}
}

@article{guo2024large,
  title={Large language model based multi-agents: A survey of progress and challenges},
  author={Guo, Taicheng and Chen, Xiuying and Wang, Yaqi and Chang, Ruidi and Pei, Shichao and Chawla, Nitesh V and Wiest, Olaf and Zhang, Xiangliang},
  journal={arXiv preprint arXiv:2402.01680},
  year={2024}
}

@article{zhang2024vision,
  title={Vision-language models for vision tasks: A survey},
  author={Zhang, Jingyi and Huang, Jiaxing and Jin, Sheng and Lu, Shijian},
  journal={IEEE transactions on pattern analysis and machine intelligence},
  volume={46},
  number={8},
  pages={5625--5644},
  year={2024},
  publisher={IEEE}
}

@article{chen2025opengpt,
  title={OpenGPT-4o-Image: A Comprehensive Dataset for Advanced Image Generation and Editing},
  author={Chen, Zhihong and Bai, Xuehai and Shi, Yang and Fu, Chaoyou and Zhang, Huanyu and Wang, Haotian and Sun, Xiaoyan and Zhang, Zhang and Wang, Liang and Zhang, Yuanxing and others},
  journal={arXiv preprint arXiv:2509.24900},
  year={2025}
}

@article{lecun1998mnist,
  title={The MNIST database of handwritten digits},
  author={LeCun, Yann},
  journal={http://yann. lecun. com/exdb/mnist/},
  year={1998}
}

@article{voulodimos2018deep,
  title={Deep learning for computer vision: A brief review},
  author={Voulodimos, Athanasios and Doulamis, Nikolaos and Doulamis, Anastasios and Protopapadakis, Eftychios},
  journal={Computational intelligence and neuroscience},
  volume={2018},
  number={1},
  pages={7068349},
  year={2018},
  publisher={Wiley Online Library}
}

@book{szeliski2022computer,
  title={Computer vision: algorithms and applications},
  author={Szeliski, Richard},
  year={2022},
  publisher={Springer Nature}
}

@book{forsyth2002computer,
  title={Computer vision: a modern approach},
  author={Forsyth, David A and Ponce, Jean},
  year={2002},
  publisher={prentice hall professional technical reference}
}

@article{zhao2024review,
  title={A review of convolutional neural networks in computer vision},
  author={Zhao, Xia and Wang, Limin and Zhang, Yufei and Han, Xuming and Deveci, Muhammet and Parmar, Milan},
  journal={Artificial Intelligence Review},
  volume={57},
  number={4},
  pages={99},
  year={2024},
  publisher={Springer}
}

@article{zhang2024deep,
  title={Deep Loss Convexification for Learning Iterative Models},
  author={Zhang, Ziming and Shao, Yuping and Zhang, Yiqing and Lin, Fangzhou and Zhang, Haichong and Rundensteiner, Elke},
  journal={IEEE Transactions on Pattern Analysis and Machine Intelligence},
  year={2024},
  publisher={IEEE}
}

@inproceedings{deng2009imagenet,
  title={Imagenet: A large-scale hierarchical image database},
  author={Deng, Jia and Dong, Wei and Socher, Richard and Li, Li-Jia and Li, Kai and Fei-Fei, Li},
  booktitle={2009 IEEE conference on computer vision and pattern recognition},
  pages={248--255},
  year={2009},
  organization={Ieee}
}

@article{elngar2021image,
  title={Image classification based on CNN: a survey},
  author={Elngar, Ahmed A and Arafa, Mohamed and Fathy, Amar and Moustafa, Basma and Mahmoud, Omar and Shaban, Mohamed and Fawzy, Nehal},
  journal={Journal of Cybersecurity and Information Management},
  volume={6},
  number={1},
  pages={18--50},
  year={2021}
}

@article{minaee2021image,
  title={Image segmentation using deep learning: A survey},
  author={Minaee, Shervin and Boykov, Yuri and Porikli, Fatih and Plaza, Antonio and Kehtarnavaz, Nasser and Terzopoulos, Demetri},
  journal={IEEE transactions on pattern analysis and machine intelligence},
  volume={44},
  number={7},
  pages={3523--3542},
  year={2021},
  publisher={IEEE}
}

@article{zou2023object,
  title={Object detection in 20 years: A survey},
  author={Zou, Zhengxia and Chen, Keyan and Shi, Zhenwei and Guo, Yuhong and Ye, Jieping},
  journal={Proceedings of the IEEE},
  volume={111},
  number={3},
  pages={257--276},
  year={2023},
  publisher={IEEE}
}

@article{theis2024makes,
  title={What makes an image realistic?},
  author={Theis, Lucas},
  journal={arXiv preprint arXiv:2403.04493},
  year={2024}
}

@article{chen2023hierarchical,
  title={Hierarchical integration diffusion model for realistic image deblurring},
  author={Chen, Zheng and Zhang, Yulun and Liu, Ding and Gu, Jinjin and Kong, Linghe and Yuan, Xin and others},
  journal={Advances in neural information processing systems},
  volume={36},
  pages={29114--29125},
  year={2023}
}

@inproceedings{li2023towards,
  title={Towards benchmarking and assessing visual naturalness of physical world adversarial attacks},
  author={Li, Simin and Zhang, Shuning and Chen, Gujun and Wang, Dong and Feng, Pu and Wang, Jiakai and Liu, Aishan and Yi, Xin and Liu, Xianglong},
  booktitle={Proceedings of the IEEE/CVF Conference on Computer Vision and Pattern Recognition},
  pages={12324--12333},
  year={2023}
}

@article{chen2023exploring,
  title={Exploring the naturalness of ai-generated images},
  author={Chen, Zijian and Sun, Wei and Wu, Haoning and Zhang, Zicheng and Jia, Jun and Ji, Zhongpeng and Sun, Fengyu and Jui, Shangling and Min, Xiongkuo and Zhai, Guangtao and others},
  journal={arXiv preprint arXiv:2312.05476},
  year={2023}
}

@article{wang2025vision,
  title={Vision-Zero: Scalable VLM Self-Improvement via Strategic Gamified Self-Play},
  author={Wang, Qinsi and Liu, Bo and Zhou, Tianyi and Shi, Jing and Lin, Yueqian and Chen, Yiran and Li, Hai Helen and Wan, Kun and Zhao, Wentian},
  journal={arXiv preprint arXiv:2509.25541},
  year={2025}
}

@article{lu2024deepseek,
  title={Deepseek-vl: towards real-world vision-language understanding},
  author={Lu, Haoyu and Liu, Wen and Zhang, Bo and Wang, Bingxuan and Dong, Kai and Liu, Bo and Sun, Jingxiang and Ren, Tongzheng and Li, Zhuoshu and Yang, Hao and others},
  journal={arXiv preprint arXiv:2403.05525},
  year={2024}
}

@article{wang2024rl,
  title={Rl-vlm-f: Reinforcement learning from vision language foundation model feedback},
  author={Wang, Yufei and Sun, Zhanyi and Zhang, Jesse and Xian, Zhou and Biyik, Erdem and Held, David and Erickson, Zackory},
  journal={arXiv preprint arXiv:2402.03681},
  year={2024}
}

@article{magicbrush,
  title={Magicbrush: A manually annotated dataset for instruction-guided image editing},
  author={Zhang, Kai and Mo, Lingbo and Chen, Wenhu and Sun, Huan and Su, Yu},
  journal={Advances in Neural Information Processing Systems},
  volume={36},
  pages={31428--31449},
  year={2023}
}

@article{bytemorph,
  title={ByteMorph: Benchmarking Instruction-Guided Image Editing with Non-Rigid Motions},
  author={Chang, Di and Cao, Mingdeng and Shi, Yichun and Liu, Bo and Cai, Shengqu and Zhou, Shijie and Huang, Weilin and Wetzstein, Gordon and Soleymani, Mohammad and Wang, Peng},
  journal={arXiv preprint arXiv:2506.03107},
  year={2025}
}

@inproceedings{referringimageediting,
  title={Referring image editing: Object-level image editing via referring expressions},
  author={Liu, Chang and Li, Xiangtai and Ding, Henghui},
  booktitle={Proceedings of the IEEE/CVF Conference on Computer Vision and Pattern Recognition},
  pages={13128--13138},
  year={2024}
}

@article{iharmony4,
  title={Image harmonization dataset iharmony4: Hcoco, hadobe5k, hflickr, and hday2night},
  author={Cong, Wenyan and Zhang, Jianfu and Niu, Li and Liu, Liu and Ling, Zhixin and Li, Weiyuan and Zhang, Liqing},
  journal={arXiv preprint arXiv:1908.10526},
  year={2019}
}

@inproceedings{orida,
  title={ORIDa: Object-centric Real-world Image Composition Dataset},
  author={Kim, Jinwoo and Han, Sangmin and Jeong, Jinho and Choi, Jiwoo and Kim, Dongyeoung and Kim, Seon Joo},
  booktitle={Proceedings of the Computer Vision and Pattern Recognition Conference},
  pages={3051--3060},
  year={2025}
}

@inproceedings{objectdrop,
  title={Objectdrop: Bootstrapping counterfactuals for photorealistic object removal and insertion},
  author={Winter, Daniel and Cohen, Matan and Fruchter, Shlomi and Pritch, Yael and Rav-Acha, Alex and Hoshen, Yedid},
  booktitle={European Conference on Computer Vision},
  pages={112--129},
  year={2024},
  organization={Springer}
}

@article{placement,
  title={Add-it: Training-free object insertion in images with pretrained diffusion models},
  author={Tewel, Yoad and Gal, Rinon and Samuel, Dvir and Atzmon, Yuval and Wolf, Lior and Chechik, Gal},
  journal={arXiv preprint arXiv:2411.07232},
  year={2024}
}

@article{comanici2025gemini,
  title={Gemini 2.5: Pushing the frontier with advanced reasoning, multimodality, long context, and next generation agentic capabilities},
  author={Comanici, Gheorghe and Bieber, Eric and Schaekermann, Mike and Pasupat, Ice and Sachdeva, Noveen and Dhillon, Inderjit and Blistein, Marcel and Ram, Ori and Zhang, Dan and Rosen, Evan and others},
  journal={arXiv preprint arXiv:2507.06261},
  year={2025}
}

@article{wu2025editreward,
  title={Editreward: A human-aligned reward model for instruction-guided image editing},
  author={Wu, Keming and Jiang, Sicong and Ku, Max and Nie, Ping and Liu, Minghao and Chen, Wenhu},
  journal={arXiv preprint arXiv:2509.26346},
  year={2025}
}

@article{lin2025uniworld,
  title={Uniworld: High-resolution semantic encoders for unified visual understanding and generation},
  author={Lin, Bin and Li, Zongjian and Cheng, Xinhua and Niu, Yuwei and Ye, Yang and He, Xianyi and Yuan, Shenghai and Yu, Wangbo and Wang, Shaodong and Ge, Yunyang and others},
  journal={arXiv preprint arXiv:2506.03147},
  year={2025}
}

@article{qian2025pico,
  title={Pico-Banana-400K: A Large-Scale Dataset for Text-Guided Image Editing},
  author={Qian, Yusu and Bocek-Rivele, Eli and Song, Liangchen and Tong, Jialing and Yang, Yinfei and Lu, Jiasen and Hu, Wenze and Gan, Zhe},
  journal={arXiv preprint arXiv:2510.19808},
  year={2025}
}

@article{ku2023imagenhub,
  title={Imagenhub: Standardizing the evaluation of conditional image generation models},
  author={Ku, Max and Li, Tianle and Zhang, Kai and Lu, Yujie and Fu, Xingyu and Zhuang, Wenwen and Chen, Wenhu},
  journal={arXiv preprint arXiv:2310.01596},
  year={2023}
}

@article{peng2024dreambench++,
  title={Dreambench++: A human-aligned benchmark for personalized image generation},
  author={Peng, Yuang and Cui, Yuxin and Tang, Haomiao and Qi, Zekun and Dong, Runpei and Bai, Jing and Han, Chunrui and Ge, Zheng and Zhang, Xiangyu and Xia, Shu-Tao},
  journal={arXiv preprint arXiv:2406.16855},
  year={2024}
}

@article{jiang2024genai,
  title={Genai arena: An open evaluation platform for generative models},
  author={Jiang, Dongfu and Ku, Max and Li, Tianle and Ni, Yuansheng and Sun, Shizhuo and Fan, Rongqi and Chen, Wenhu},
  journal={Advances in Neural Information Processing Systems},
  volume={37},
  pages={79889--79908},
  year={2024}
}

@inproceedings{zhao2025challenges,
  title={Challenges in trustworthy human evaluation of chatbots},
  author={Zhao, Wenting and Rush, Alexander M and Goyal, Tanya},
  booktitle={Findings of the Association for Computational Linguistics: NAACL 2025},
  pages={3359--3365},
  year={2025}
}

@article{instructp2p,
	title        = {InstructPix2Pix: Learning to Follow Image Editing Instructions},
	author       = {Tim Brooks and Aleksander Holynski and Alexei A. Efros},
	year         = 2022,
	journal      = {2023 IEEE/CVF Conference on Computer Vision and Pattern Recognition (CVPR)},
	pages        = {18392--18402},
	url          = {https://api.semanticscholar.org/CorpusID:253581213}
}

@article{hqedit,
	title        = {Hq-edit: A high-quality dataset for instruction-based image editing},
	author       = {Hui, Mude and Yang, Siwei and Zhao, Bingchen and Shi, Yichun and Wang, Heng and Wang, Peng and Zhou, Yuyin and Xie, Cihang},
	year         = 2024,
	journal      = {arXiv preprint arXiv:2404.09990}
}

@article{seeddataedit,
	title        = {Seed-data-edit technical report: A hybrid dataset for instructional image editing},
	author       = {Ge, Yuying and Zhao, Sijie and Li, Chen and Ge, Yixiao and Shan, Ying},
	year         = 2024,
	journal      = {arXiv preprint arXiv:2405.04007}
}

@article{ultraedit,
	title        = {UltraEdit: Instruction-based fine-grained image editing at scale},
	author       = {Zhao, Haozhe and Ma, Xiaojian Shawn and Chen, Liang and Si, Shuzheng and Wu, Rujie and An, Kaikai and Yu, Peiyu and Zhang, Minjia and Li, Qing and Chang, Baobao},
	year         = 2024,
	journal      = {Advances in Neural Information Processing Systems},
	volume       = 37,
	pages        = {3058--3093}
}

@misc{gpt-image-1,
	title        = {GPT Image 1},
	author       = {OpenAI},
	year         = 2025,
	url          = {https://platform.openai.com/docs/models/gpt-image-1}
}

@misc{gpt-image-1_5,
	title        = {The new ChatGPT Images is here},
	author       = {OpenAI},
	year         = 2025,
	url          = {https://openai.com/index/new-chatgpt-images-is-here/}
}

@misc{nano-banana,
	title        = {Introducing Gemini 2.5 Flash Image, our state-of-the-art image model},
	author       = {Google DeepMind},
	year         = 2025,
	url          = {https://developers.googleblog.com/introducing-gemini-2-5-flash-image/}
}

@misc{nano-banana-pro,
	title        = {Introducing Nano Banana Pro},
	author       = {Google DeepMind},
	year         = 2025,
	url          = {https://blog.google/technology/ai/nano-banana-pro/}
}

@article{ye2025imgedit,
  title={Imgedit: A unified image editing dataset and benchmark},
  author={Ye, Yang and He, Xianyi and Li, Zongjian and Lin, Bin and Yuan, Shenghai and Yan, Zhiyuan and Hou, Bohan and Yuan, Li},
  journal={arXiv preprint arXiv:2505.20275},
  year={2025}
}

@article{wan2025,
      title={Wan: Open and Advanced Large-Scale Video Generative Models}, 
      author={Team Wan and Ang Wang and Baole Ai and Bin Wen and Chaojie Mao and Chen-Wei Xie and Di Chen and Feiwu Yu and Haiming Zhao and Jianxiao Yang and Jianyuan Zeng and Jiayu Wang and Jingfeng Zhang and Jingren Zhou and Jinkai Wang and Jixuan Chen and Kai Zhu and Kang Zhao and Keyu Yan and Lianghua Huang and Mengyang Feng and Ningyi Zhang and Pandeng Li and Pingyu Wu and Ruihang Chu and Ruili Feng and Shiwei Zhang and Siyang Sun and Tao Fang and Tianxing Wang and Tianyi Gui and Tingyu Weng and Tong Shen and Wei Lin and Wei Wang and Wei Wang and Wenmeng Zhou and Wente Wang and Wenting Shen and Wenyuan Yu and Xianzhong Shi and Xiaoming Huang and Xin Xu and Yan Kou and Yangyu Lv and Yifei Li and Yijing Liu and Yiming Wang and Yingya Zhang and Yitong Huang and Yong Li and You Wu and Yu Liu and Yulin Pan and Yun Zheng and Yuntao Hong and Yupeng Shi and Yutong Feng and Zeyinzi Jiang and Zhen Han and Zhi-Fan Wu and Ziyu Liu},
      journal = {arXiv preprint arXiv:2503.20314},
      year={2025}
}

@article{varedit2025,
  title={Visual Autoregressive Modeling for Instruction-Guided Image Editing},
  author={Mao, Qingyang and Cai, Qi and Li, Yehao and Pan, Yingwei and Cheng, Mingyue and Yao, Ting and Liu, Qi and Mei, Tao},
  journal={arXiv preprint},
  year={2025}
}

@inproceedings{zhang2025icedit,
  title     = {In-Context Edit: Enabling Instructional Image Editing with In-Context Generation in Large-Scale Diffusion Transformers},
  author    = {Zhang, Zechuan and Xie, Ji and Lu, Yu and Yang, Zongxin and Yang, Yi},
  booktitle = {Advances in Neural Information Processing Systems (NeurIPS)},
  year      = {2025},
  note      = {arXiv:2504.20690}
}

@inproceedings{yu2025anyedit,
  title={Anyedit: Mastering unified high-quality image editing for any idea},
  author={Yu, Qifan and Chow, Wei and Yue, Zhongqi and Pan, Kaihang and Wu, Yang and Wan, Xiaoyang and Li, Juncheng and Tang, Siliang and Zhang, Hanwang and Zhuang, Yueting},
  booktitle={Proceedings of the Computer Vision and Pattern Recognition Conference},
  pages={26125--26135},
  year={2025}
}

@misc{chen2025instructclipimprovinginstructionguidedimage,
      title={Instruct-CLIP: Improving Instruction-Guided Image Editing with Automated Data Refinement Using Contrastive Learning}, 
      author={Sherry X. Chen and Misha Sra and Pradeep Sen},
      year={2025},
      eprint={2503.18406},
      archivePrefix={arXiv},
      primaryClass={cs.CV},
      url={https://arxiv.org/abs/2503.18406}, 
}

@article{liu2025step1x-edit,
      title={Step1X-Edit: A Practical Framework for General Image Editing}, 
      author={Shiyu Liu and Yucheng Han and Peng Xing and Fukun Yin and Rui Wang and Wei Cheng and Jiaqi Liao and Yingming Wang and Honghao Fu and Chunrui Han and Guopeng Li and Yuang Peng and Quan Sun and Jingwei Wu and Yan Cai and Zheng Ge and Ranchen Ming and Lei Xia and Xianfang Zeng and Yibo Zhu and Binxing Jiao and Xiangyu Zhang and Gang Yu and Daxin Jiang},
      journal={arXiv preprint arXiv:2504.17761},
      year={2025}
}

@misc{wu2025qwenimagetechnicalreport,
      title={Qwen-Image Technical Report}, 
      author={Chenfei Wu and Jiahao Li and Jingren Zhou and Junyang Lin and Kaiyuan Gao and Kun Yan and Sheng-ming Yin and Shuai Bai and Xiao Xu and Yilei Chen and Yuxiang Chen and Zecheng Tang and Zekai Zhang and Zhengyi Wang and An Yang and Bowen Yu and Chen Cheng and Dayiheng Liu and Deqing Li and Hang Zhang and Hao Meng and Hu Wei and Jingyuan Ni and Kai Chen and Kuan Cao and Liang Peng and Lin Qu and Minggang Wu and Peng Wang and Shuting Yu and Tingkun Wen and Wensen Feng and Xiaoxiao Xu and Yi Wang and Yichang Zhang and Yongqiang Zhu and Yujia Wu and Yuxuan Cai and Zenan Liu},
      year={2025},
      eprint={2508.02324},
      archivePrefix={arXiv},
      primaryClass={cs.CV},
      url={https://arxiv.org/abs/2508.02324}, 
}

@article{chang2025bytemorph,
  title={ByteMorph: Benchmarking Instruction-Guided Image Editing with Non-Rigid Motions},
  author={Chang, Di and Cao, Mingdeng and Shi, Yichun and Liu, Bo and Cai, Shengqu and Zhou, Shijie and Huang, Weilin and Wetzstein, Gordon and Soleymani, Mohammad and Wang, Peng},
  journal={arXiv preprint arXiv:2506.03107},
  year={2025}
}

@Article{SuperEdit,
      title={SuperEdit: Rectifying and Facilitating Supervision for Instruction-Based Image Editing},
      author={Ming Li and Xin Gu and Fan Chen and Xiaoying Xing and Longyin Wen and Chen Chen and Sijie Zhu},
      year={2025},
      eprint={2505.02370},
      archivePrefix={arXiv},
      primaryClass={cs.CV},
      url={https://arxiv.org/abs/2505.02370},
}

@inproceedings{MultiReward,
  title={Multi-Reward as Condition for Instruction-based Image Editing},
  author={Gu, Xin and Li, Ming and Zhang, Libo and Chen, Fan and Wen, Longyin and Luo, Tiejian and Zhu, Sijie},
  booktitle={The Thirteenth International Conference on Learning Representations},
  year={2025},
}

@inproceedings{qu2023exploring,
  title={Exploring stroke-level modifications for scene text editing},
  author={Qu, Yadong and Tan, Qingfeng and Xie, Hongtao and Xu, Jianjun and Wang, Yuxin and Zhang, Yongdong},
  booktitle={Proceedings of the AAAI Conference on Artificial Intelligence},
  volume={37},
  number={2},
  pages={2119--2127},
  year={2023}
}

@inproceedings{fang2025recognition,
  title={Recognition-Synergistic Scene Text Editing},
  author={Fang, Zhengyao and Lyu, Pengyuan and Wu, Jingjing and Zhang, Chengquan and Yu, Jun and Lu, Guangming and Pei, Wenjie},
  booktitle={Proceedings of the Computer Vision and Pattern Recognition Conference},
  pages={13104--13113},
  year={2025}
}

@article{chiang2024chatbot,
  title={Chatbot arena: An open platform for evaluating llms by human preference},
  author={Chiang, Wei-Lin and Zheng, Lianmin and Sheng, Ying and Angelopoulos, Anastasios Nikolas and Li, Tianle and Li, Dacheng and Zhang, Hao and Zhu, Banghua and Jordan, Michael and Gonzalez, Joseph E and others},
  journal={arXiv preprint arXiv:2403.04132},
  year={2024}
}

@article{bradley1952rank,
  title={Rank analysis of incomplete block designs: I. The method of paired comparisons},
  author={Bradley, Ralph Allan and Terry, Milton E},
  journal={Biometrika},
  volume={39},
  number={3/4},
  pages={324--345},
  year={1952},
  publisher={JSTOR}
}

@inproceedings{clip,
	title        = {Learning transferable visual models from natural language supervision},
	author       = {Radford, Alec and Kim, Jong Wook and Hallacy, Chris and Ramesh, Aditya and Goh, Gabriel and Agarwal, Sandhini and Sastry, Girish and Askell, Amanda and Mishkin, Pamela and Clark, Jack and others},
	year         = 2021,
	booktitle    = {International conference on machine learning},
	pages        = {8748--8763},
	organization = {PmLR}
}

@inproceedings{psnr,
  author={Korhonen, Jari and You, Junyong},
  booktitle={2012 Fourth International Workshop on Quality of Multimedia Experience}, 
  title={Peak signal-to-noise ratio revisited: Is simple beautiful?}, 
  year={2012},
  volume={},
  number={},
  pages={37-38},
  doi={10.1109/QoMEX.2012.6263880}}

@article{ssim,
  author={Zhou Wang and Bovik, A.C. and Sheikh, H.R. and Simoncelli, E.P.},
  journal={IEEE Transactions on Image Processing}, 
  title={Image quality assessment: from error visibility to structural similarity}, 
  year={2004},
  volume={13},
  number={4},
  pages={600-612},
  doi={10.1109/TIP.2003.819861}}

@article{seedream2025seedream,
  title={Seedream 4.0: Toward next-generation multimodal image generation},
  author={Seedream, Team and Chen, Yunpeng and Gao, Yu and Gong, Lixue and Guo, Meng and Guo, Qiushan and Guo, Zhiyao and Hou, Xiaoxia and Huang, Weilin and Huang, Yixuan and others},
  journal={arXiv preprint arXiv:2509.20427},
  year={2025}
}

@misc{flux-2-2025,
    author={Black Forest Labs},
    title={{FLUX.2: Frontier Visual Intelligence}},
    year={2025},
    howpublished={\url{https://bfl.ai/blog/flux-2}},
}

@misc{manus_1_6,
	title        = {Introducing Manus 1.6: Max Performance, Mobile Dev, and Design View},
	author       = {Manus Meta},
	year         = 2025,
	url          = {https://manus.im/blog/manus-max-release}
}

@misc{google2025gemini25card,
  title={Gemini 2.5 Pro Model Card},
  author={Google},
  year={2025},
  url={https://modelcards.withgoogle.com/assets/documents/gemini-2.5-pro.pdf},
  note={Accessed: 2026-01-11}
}

@article{lin2025jarvisevo,
  title={JarvisEvo: Towards a Self-Evolving Photo Editing Agent with Synergistic Editor-Evaluator Optimization},
  author={Lin, Yunlong and Wang, Linqing and Lin, Kunjie and Lin, Zixu and Gong, Kaixiong and Li, Wenbo and Lin, Bin and Li, Zhenxi and Zhang, Shiyi and Peng, Yuyang and others},
  journal={arXiv preprint arXiv:2511.23002},
  year={2025}
}

@misc{OpenAI2025ChatGPTagent,
  title={Introducing ChatGPT agent: bridging research and action},
  author={OpenAI},
  year={2025},
  url={https://openai.com/index/introducing-chatgpt-agent/},
  note={Accessed: 2026-01-26}
}

@inproceedings{yao2022react,
  title={React: Synergizing reasoning and acting in language models},
  author={Yao, Shunyu and Zhao, Jeffrey and Yu, Dian and Du, Nan and Shafran, Izhak and Narasimhan, Karthik R and Cao, Yuan},
  booktitle={The eleventh international conference on learning representations},
  year={2022}
}

@article{zheng2023judging,
  title={Judging llm-as-a-judge with mt-bench and chatbot arena},
  author={Zheng, Lianmin and Chiang, Wei-Lin and Sheng, Ying and Zhuang, Siyuan and Wu, Zhanghao and Zhuang, Yonghao and Lin, Zi and Li, Zhuohan and Li, Dacheng and Xing, Eric and others},
  journal={Advances in neural information processing systems},
  volume={36},
  pages={46595--46623},
  year={2023}
}

@article{ouyang2022training,
  title={Training language models to follow instructions with human feedback},
  author={Ouyang, Long and Wu, Jeffrey and Jiang, Xu and Almeida, Diogo and Wainwright, Carroll and Mishkin, Pamela and Zhang, Chong and Agarwal, Sandhini and Slama, Katarina and Ray, Alex and others},
  journal={Advances in neural information processing systems},
  volume={35},
  pages={27730--27744},
  year={2022}
}

@article{kendall1938new,
  title={A new measure of rank correlation},
  author={Kendall, Maurice G},
  journal={Biometrika},
  volume={30},
  number={1-2},
  pages={81--93},
  year={1938},
  publisher={Oxford University Press}
}

@article{dubois2024length,
  title={Length-controlled alpacaeval: A simple way to debias automatic evaluators},
  author={Dubois, Yann and Galambosi, Bal{\'a}zs and Liang, Percy and Hashimoto, Tatsunori B},
  journal={arXiv preprint arXiv:2404.04475},
  year={2024}
}

@article{dubois2023alpacafarm,
  title={Alpacafarm: A simulation framework for methods that learn from human feedback},
  author={Dubois, Yann and Li, Chen Xuechen and Taori, Rohan and Zhang, Tianyi and Gulrajani, Ishaan and Ba, Jimmy and Guestrin, Carlos and Liang, Percy S and Hashimoto, Tatsunori B},
  journal={Advances in Neural Information Processing Systems},
  volume={36},
  pages={30039--30069},
  year={2023}
}

@inproceedings{luo2025visual,
  title={Visual-instructed degradation diffusion for all-in-one image restoration},
  author={Luo, Wenyang and Qin, Haina and Chen, Zewen and Wang, Libin and Zheng, Dandan and Li, Yuming and Liu, Yufan and Li, Bing and Hu, Weiming},
  booktitle={Proceedings of the Computer Vision and Pattern Recognition Conference},
  pages={12764--12777},
  year={2025}
}

@inproceedings{janjua2026grounding,
  title={Grounding degradations in natural language for all-in-one video restoration},
  author={Janjua, Muhammad Kamran and Ghasemabadi, Amirhosein and Zhang, Kunlin and Salameh, Mohammad and Gao, Chao and Niu, Di},
  booktitle={Proceedings of the IEEE/CVF Winter Conference on Applications of Computer Vision},
  pages={5734--5743},
  year={2026}
}

@inproceedings{wang2025adapting,
  title={Adapting text-to-image generation with feature difference instruction for generic image restoration},
  author={Wang, Chao and Fan, Hehe and Yang, Huichen and Karimi, Sarvnaz and Yao, Lina and Yang, Yi},
  booktitle={Proceedings of the Computer Vision and Pattern Recognition Conference},
  pages={23539--23550},
  year={2025}
}

@inproceedings{zhou2025low,
  title={Low-light image enhancement via generative perceptual priors},
  author={Zhou, Han and Dong, Wei and Liu, Xiaohong and Zhang, Yulun and Zhai, Guangtao and Chen, Jun},
  booktitle={Proceedings of the AAAI Conference on Artificial Intelligence},
  volume={39},
  number={10},
  pages={10752--10760},
  year={2025}
}

@inproceedings{gu2025improving,
  title={Improving Visual and Downstream Performance of Low-Light Enhancer with Vision Foundation Models Collaboration},
  author={Gu, Yuxuan and Wang, Haoxuan and Ling, Pengyang and Wei, Zhixiang and Chen, Huaian and Jin, Yi and Chen, Enhong},
  booktitle={Proceedings of the Computer Vision and Pattern Recognition Conference},
  pages={16071--16080},
  year={2025}
}

@inproceedings{zhang2026adaptive,
  title={Adaptive Dynamic Dehazing via Instruction-Driven and Task-Feedback Closed-Loop Optimization for Diverse Downstream Task Adaptation},
  author={Zhang, Yafei and Song, Shuaitian and Li, Huafeng and Wang, Shujuan and Liu, Yu},
  booktitle={Proceedings of the AAAI Conference on Artificial Intelligence},
  volume={40},
  number={15},
  pages={12888--12896},
  year={2026}
}

@inproceedings{cao2025instruction,
  title={Instruction-based image manipulation by watching how things move},
  author={Cao, Mingdeng and Zhang, Xuaner and Zheng, Yinqiang and Xia, Zhihao},
  booktitle={Proceedings of the Computer Vision and Pattern Recognition Conference},
  pages={2704--2713},
  year={2025}
}

@inproceedings{song2026insert,
  title={Insert anything: Image insertion via in-context editing in dit},
  author={Song, Wensong and Jiang, Hong and Yang, Zongxin and Cheng, Zheqiao and Quan, Ruijie and Yang, Yi},
  booktitle={Proceedings of the AAAI Conference on Artificial Intelligence},
  volume={40},
  number={11},
  pages={9097--9105},
  year={2026}
}

@article{jia2025compbench,
  title={Compbench: Benchmarking complex instruction-guided image editing},
  author={Jia, Bohan and Huang, Wenxuan and Tang, Yuntian and Qiao, Junbo and Liao, Jincheng and Cao, Shaosheng and Zhao, Fei and Feng, Zhaopeng and Gu, Zhouhong and Yin, Zhenfei and others},
  journal={arXiv preprint arXiv:2505.12200},
  year={2025}
}

@article{zhang2025self,
  title={Self-prompt guided image outpainting model for captions absence in social scenes},
  author={Zhang, Zongyan and Chen, CL Philip and Weng, Haohan and Zhang, Tong},
  journal={IEEE Transactions on Computational Social Systems},
  year={2025},
  publisher={IEEE}
}

@inproceedings{wang2025complexbench,
  title={Complexbench-edit: Benchmarking complex instruction-driven image editing via compositional dependencies},
  author={Wang, Chenglin and Zhou, Yucheng and Wang, Qianning and Wang, Zhe and Zhang, Kai},
  booktitle={Proceedings of the 33rd ACM International Conference on Multimedia},
  pages={13391--13397},
  year={2025}
}

@article{hafner2021clip,
  title={CLIP and complementary methods},
  author={Hafner, Markus and Katsantoni, Maria and K{\"o}ster, Tino and Marks, James and Mukherjee, Joyita and Staiger, Dorothee and Ule, Jernej and Zavolan, Mihaela},
  journal={Nature Reviews Methods Primers},
  volume={1},
  number={1},
  pages={20},
  year={2021},
  publisher={Nature Publishing Group UK London}
}

@article{zhang2022dino,
  title={Dino: Detr with improved denoising anchor boxes for end-to-end object detection},
  author={Zhang, Hao and Li, Feng and Liu, Shilong and Zhang, Lei and Su, Hang and Zhu, Jun and Ni, Lionel M and Shum, Heung-Yeung},
  journal={arXiv preprint arXiv:2203.03605},
  year={2022}
}

@inproceedings{hessel2021clipscore,
  title={Clipscore: A reference-free evaluation metric for image captioning},
  author={Hessel, Jack and Holtzman, Ari and Forbes, Maxwell and Le Bras, Ronan and Choi, Yejin},
  booktitle={Proceedings of the 2021 conference on empirical methods in natural language processing},
  pages={7514--7528},
  year={2021}
}
